\documentclass[10pt,twocolumn,letterpaper]{article}

\usepackage{cvpr}      
\usepackage{times}
\usepackage{epsfig}
\usepackage{graphicx}
\usepackage{amsmath}
\usepackage{amssymb}
\usepackage{color}
\usepackage{pifont}
\usepackage{booktabs}
\usepackage{multirow}
\newcommand{\tabincell}[2]{\begin{tabular}{@{}#1@{}}#2\end{tabular}}

\usepackage[hang,flushmargin]{footmisc}

\usepackage[pagebackref=true,breaklinks=true,letterpaper=true,colorlinks,bookmarks=false]{hyperref}

\def\cvprPaperID{92} 
\def\confName{CVPR}
\def\confYear{2022}

\begin{document}
	
	\title{NTIRE 2022 Challenge on Stereo Image Super-Resolution: Methods and Results}
	\author{Longguang Wang, ~~Yulan Guo*, ~~Yingqian Wang, ~~Juncheng Li, ~~Shuhang Gu, ~~Radu Timofte,\\
	    Liangyu Chen, ~~Xiaojie Chu, ~~Wenqing Yu, ~~Kai Jin, ~~Zeqiang Wei, ~~Sha Guo, ~~Angulia Yang,\\ Xiuzhuang Zhou, ~~Guodong Guo, ~~Bin Dai, ~~Feiyue Peng, ~~Huaxin Xiao, ~~Shen Yan, ~~Yuxiang Liu,\\
	    Hanxiao Cai, ~~Pu Cao, ~~Yang Nie, ~~Lu Yang, ~~Qing Song, ~~Xiaotao Hu, ~~Jun Xu, ~~Mai Xu, \\
	    Junpeng Jing, ~~Xin Deng, ~~Qunliang Xing, ~~Minglang Qiao, ~~Zhenyu Guan, ~~Wenlong Guo,\\
	    Chenxu Peng, ~~Zan Chen, ~~Junyang Chen, ~~Hao Li, ~~Junbin Chen, ~~Weijie Li, ~~Zhijing Yang,\\
	    Gen Li, ~~Aijin Li, ~~Lei Sun, ~~Dafeng Zhang, ~~Shizhuo Liu, ~~Jiangtao Zhang, ~~Yanyun Qu,\\
	    Hao-Hsiang Yang, ~~Zhi-Kai Huang, ~~Wei-Ting Chen, ~~Hua-En Chang, ~~Sy-Yen Kuo,\\
	    Qiaohui Liang, ~~Jianxin Lin, ~~Yijun Wang, ~~Lianying Yin, ~~Rongju Zhang, ~~Wei Zhao,\\
	    Peng Xiao, ~~Rongjian Xu, ~~Zhilu Zhang, ~~Wangmeng Zuo, ~~Hansheng Guo, ~~Guangwei Gao,\\
	    Tieyong Zeng, ~~Huicheng Pi, ~~Shunli Zhang, ~~Joohyeok Kim, ~~HyeonA Kim, ~~Eunpil Park,\\
	    Jae-Young Sim, ~~Jucai Zhai, ~~Pengcheng Zeng, ~~Yang Liu, ~~Chihao Ma, ~~Yulin Huang, ~~Junying Chen 
	}
	
	
	\maketitle

	\begin{abstract}
		In this paper, we summarize the 1st NTIRE challenge on stereo image super-resolution (restoration of rich details in a pair of low-resolution stereo images) with a focus on new solutions and results. This challenge has 1 track aiming at the stereo image super-resolution problem under a standard bicubic degradation. In total, 238  participants were successfully registered, and 21 teams competed in the final testing phase. Among those participants, 20 teams successfully submitted results with PSNR (RGB) scores better than the baseline. This challenge establishes a new benchmark for stereo image SR. 
	\end{abstract}
	
	\section{Introduction}
	\footnotetext{
	\noindent *Corresponding author: Yulan Guo (yulan.guo@nudt.edu.cn).	
	\\~~Section \ref{appendix} provides the authors and affiliations of each team.
	\\~~NTIRE 2022 webpage: \url{https://data.vision.ee.ethz.ch/cvl/ntire22/}
	\\~~Challenge webpage: \url{https://codalab.lisn.upsaclay.fr/competitions/1598}
	\\~~Leaderboard: \url{https://codalab.lisn.upsaclay.fr/competitions/1598\#results}
	\\~~Github: \url{https://github.com/The-Learning-And-Vision-Atelier-LAVA/Stereo-Image-SR/tree/NTIRE2022}
	}
	Stereo image pairs can encode 3D scene cues into stereo correspondences between the left and right images. 
    With the popularity of dual cameras in mobile phones, autonomous vehicles and robots, stereo vision has attracted increasingly attention in both academia and industry. 
    In many applications like AR/VR,  and robot navigation, increasing the resolution of stereo images is highly demanded to achieve higher perceptual quality and help to parse the real world.

    In recent years, remarkable progress of image super-resolution (SR) have been witnessed with deep learning techniques. {Most existing approaches focus on super-resolving single images. However, these methods cannot make full use of the cross-view information in stereo images.} Recent CNN-based video SR methods incorporate optical flow estimation and SR in unified networks to exploit temporal information in multiple frames. Nevertheless, these methods usually suffer limited performance on stereo image SR since the disparity can be much larger than their receptive fields.
    
    Stereo image SR aims to reconstruct a pair of high-resolution (HR) stereo images from a pair of low-resolution (LR) observations. Since disparities between stereo images can vary significantly for different baselines, focal lengths, depths and resolutions, it is highly challenging to incorporate stereo correspondence for stereo image SR.
    
    The NTIRE 2022 stereo image SR challenge takes a step forward to establish a benchmark for stereo image SR. It uses the Flick1024 dataset \cite{Wang2019Flickr1024} and employs standard bicubic degradation. 
    
    This challenge is one of the NTIRE 2022 associated challenges: spectral recovery~\cite{arad2022ntirerecovery}, spectral demosaicing~\cite{arad2022ntiredemosaicing},
    perceptual image quality assessment~\cite{gu2022ntire},
    inpainting~\cite{romero2022ntire},
    night photography rendering~\cite{ershov2022ntire},
    efficient super-resolution~\cite{li2022ntire},
    learning the super-resolution space~\cite{lugmayr2022ntire},
    super-resolution and quality enhancement of compressed video~\cite{yang2022ntire},
    high dynamic range~\cite{perezpellitero2022ntire},
    stereo image super-resolution,
    and burst super-resolution~\cite{bhat2022ntire}.

    \section{Related Work}\label{SecRelatedWork}
    In this section, we briefly review several major works on single image and stereo image SR.

    \subsection{Single Image SR}
    Single image SR is a long-standing problem and has been investigated for decades. In the past 10 years, deep learning-based single image SR methods have achieved promising performance. 
    
    Dong et al. \cite{Dong2014Learning} proposed the first CNN-based SR network called SRCNN to reconstruct HR images from LR inputs. Kim et al. \cite{Kim2016Accurate} proposed a deeper network with 20 layers (\textit{i.e.}, VDSR) to improve SR performance. Afterwards, SR networks became increasingly deep and complex, and thus more powerful in intra-view information exploitation. Lim et al. \cite{Lim2017Enhanced} proposed an enhanced deep SR network (\textit{i.e.}, EDSR) using both local and residual connections. Zhang et al. \cite{Zhang2018Residual} combined residual connection \cite{He2016Deep} with dense connection \cite{Huang2017Densely}, and proposed residual dense network (\textit{i.e.}, RDN) to fully use hierarchical feature representations for image SR. Subsequently, Zhang et al. \cite{Zhang2020Residual} further improved SR performance by designing a residual-in-residual network with channel attention. Li et al. \cite{li2018multi} suggested to make full use of image features with different scales and proposed a multi-scale residual network (\textit{i.e.}, MSRN). Recently, Transformer has been widely used in computer vision and achieved promising performance. In the area of low-level vision, Liang et al. \cite{liang2021swinir} applied Swin Transformer \cite{Liu2021Swin} to image restoration, and designed a SwinIR network to achieve state-of-the-art performance on single image SR. Lu et al. \cite{lu2021efficient} proposed an effective super-resolution Transformer (\textit{i.e.}, ESRT) for SISR, which reduces GPU memory consumption through a lightweight Transformer and feature separation strategy. Readers can refer to recent surveys \cite{Wang2019Deep,li2021beginner,liu2021blind} to learn more details about single image SR.
    
    
    \subsection{Stereo Image SR}
    Compared to single image SR in which only context information within one view is available, stereo image SR can use the additional information provided by the second view (\textit{i.e.}, cross-view information) to improve SR performance. However, since an object is projected onto different locations in a stereo image pair, the cross-view information is hindered to be fully exploited.
    
    To handle this disparity issue, Jeon et al. \cite{Jeon2018Enhancing} proposed a network (\textit{i.e.}, StereoSR) to learn a parallax prior by jointly training two cascaded sub-networks. The cross-view information is integrated by concatenating the left image and a stack of right images with different pre-defined shifts. Wang et al. \cite{Wang2019Learning,Wang2020Parallax} proposed a parallax attention module (PAM) to model stereo correspondence with a global receptive field along the epipolar line. The proposed PASSRnet achieves better performance than StereoSR and is more flexible with disparity variation. Based on parallax attention mechanism, Ying et al. \cite{ying2020stereo} proposed a stereo attention module and embedded it into pre-trained SISR networks for stereo image SR. Song et al. \cite{Song2020Stereoscopic} combined self-attention with parallax attention and proposed a SPAMnet for stereo image SR. Yan et al. \cite{Yan2020Disparity} proposed a domain adaptive stereo SR network (DASSR) in which the disparity was firstly estimated by using a pretrained stereo matching network and the views were warped to the other side to incorporate cross-view information. Xu et al. \cite{xu2021deep} incorporated the idea of bilateral grid processing in a CNN framework and proposed a bilateral stereo SR network.
    
    More recently, Wang et al. \cite{Wang2021Symmetric} modified PAM \cite{Wang2019Learning} to be bidirectional and symmetric, and developed an improved version of PASSRnet (\textit{i.e.}, iPASSR) to handle a series of practical issues (\textit{e.g.}, illuminance variation and occlusions) in stereo image SR. Dai et al. \cite{dai2021feedback} proposed a feedback network to alternately solve disparity estimation and stereo image SR in a recurrent manner. Ma et al. \cite{ma2021perception} proposed a GAN-based perception-oriented stereo image SR method that can generate visually pleasing and stereo consistent details. Xu et al. \cite{xu2021stereo} tackled the stereo video SR problem by simultaneously utilizing both cross-view and temporal information. 
    
	\section{NTIRE 2022 Challenge}
	The objectives of the NTIRE 2022 challenge on example-based stereo image SR are: (i) to gauge and push the state-of-the-art in SR; and (ii) to compare different solutions.
	
	\subsection{Dataset}
	The Flickr1024 dataset \cite{Wang2019Flickr1024} is used in the challenge. Flickr1024 has 1024 pairs of  RGB images with 800 for training, 112 for validation and 112 for testing purposes. The manually collected high quality images in Flickr1024 have diverse contents and rich details. In this challenge, we use Flickr1024 for both training and validation, and collect another 100 LR stereo image pairs (with private groundtruth HR images) for test.
	
	\subsection{Track and Competition}
	\noindent{\textbf{Track: Bicubic degradation.}} Standard bicubic degradation (Matlab \textit{imresize} function with default settings) is used to synthesize LR stereo images from HR ones for both training, validation and test sets.
	
	\noindent{\textbf{Challenge phases}}\\
	\noindent\textbf{(1)} \textit{\textbf{Development phase:}} 
	The participants were provided with pairs of LR and HR training images and LR validation images of the Flickr1024 dataset. The participants had the opportunity to test their solutions on the LR validation images and to receive immediate feedback by uploading their results to the server. A validation leaderboard is available online.
	
	\noindent\textbf{(2)} \textit{\textbf{Testing phase:}} 
	The participants were provided with the LR test images and were asked to submit their super-resolved images, codes, and a fact sheet for their methods before the challenge deadline. After the end of the challenge, the final results were released to the participants.
	
	\noindent{\textbf{Evaluation protocol.}} 
	The quantitative metrics are Peak signal-to-noise ratio (PSNR) in deciBels [dB] and the structural similarity index (SSIM). These full-reference measures are calculated in the RGB and Y (luminance) channels, respectively. Results are averaged over all images (for both left and right images).
	
	\section{Challenge Results}
	Among the 238 registered participants, 21 teams successfully participated the final phase and submitted their results, codes, and factsheets. Table \ref{tab1} reports the final test results, rankings of the challenge, and major details from the factsheets of 20 teams with PSNR (RGB) scores outperforming the baseline. These methods are briefly described in Section \ref{sec4} and the team members are listed in Appendix \ref{appendix}.

	\noindent \textbf{Architectures and main ideas.}
	All the proposed methods are based on deep learning techniques. Transformers (particularly SwinIR) are used in 16 solutions as the basic architecture. To exploit cross-view information, parallax-attention mechanism (PAM) are adopted in 14 solutions to capture stereo correspondence.
	
	\noindent \textbf{Restoration fidelity.}
	The top 2 methods, (\textit{i.e.}, The Fat, The Thin and The Young team and the BigoSR team), achieved similar PSNR scores (with a difference less than 0.08dB). The BUAA-MC2 entry, which ranks $6^{th}$, is only 0.21dB behind the best PSNR score of The Fat, The Thin and The Young team.
    
    \noindent \textbf{Data Augmentation.}
    Widely applied data augmentation approaches such as random flipping are used for most solutions. In addition, random horizontal shifting, random RGB channel shuffling and Cutblur~\cite{yoo2020rethinking} are also used in several solutions and help to achieve superior performance.
    
	\noindent \textbf{Ensembles and fusion.}
	Ensemble strategy (including both data ensemble and model ensemble) is adopted in several solutions to further boost the final SR performance. For data ensemble, the inputs are flipped and the resultant SR results are aligned and averaged for enhanced prediction. For model ensemble, the results produced by multiple models are averaged for better results.

	
	\noindent \textbf{Conclusions.}
	By analyzing the settings, the proposed methods and their results, we can conclude that: 
	1) The proposed methods improve the state-of-the-art in stereo image SR.
	2) Transformers are increasingly popular in stereo image SR tasks and produce significant performance improvements over CNNs.
	3) Cross-view information lying at varying disparities is critical to the stereo image SR task and helps to achieve higher performance.
    4) Benefited from bags of tricks including delicate data augmentation strategies, several single image SR solutions also produces competitive results.

	\begin{table*}[t]
		\caption{NTIRE 2022 Stereo Image SR Challenge results, final rankings, and details from the factsheets. Note that, PSNR (RGB) is used for the final ranking. ``Transf'' denotes Transformer, ``PAM'' denotes parallax attention mechanism, and ``DConv'' represents deformable convolutions.}
		\label{tab1}
		\centering
		\scriptsize
		\setlength{\tabcolsep}{1.5mm}{
			\begin{tabular}{cllcccccccc}
				\hline
				Rank &Team
				& Authors
				& PSNR (RGB)
				& PSNR (Y)
				& SSIM (RGB)
				& SSIM (Y)
				& Transf?
				& Disparity
				& Ensemble
				\tabularnewline
				\hline
				1 & \tabincell{l}{The Fat, The Thin\\and The Young} & \tabincell{l}{L. Chen, X. Chu, W. Yu} & 23.7873 & 25.2033 & 0.7360 & 0.7438  & \ding{55} & PAM & Data \& Model
				\tabularnewline
				2 & BigoSR & \tabincell{l}{K. Jin, Z. Wei, S. Guo, et al.} & 23.7126 & 25.1305 & 0.7295 & 0.7379 & \ding{51} & PAM & Data \& Model
				\tabularnewline
				3 & NUDT-CV\&CPLab & \tabincell{l}{B. Dai, F. Peng, H. Xiao, et al.} & 23.6007 & 25.0166 & 0.7287 & 0.7366 &\ding{51} & PAM &\ding{55}
				\tabularnewline
				4 & BUPT-PRIV & \tabincell{l}{P. Cao, Y. Nie, L. Yang, Q. Song} & 23.5983 & 25.0100 & 0.7217 & 0.7296 &\ding{51} & \ding{55} & Data \& Model
				\tabularnewline
				5 & NKU\_caroline & \tabincell{l}{X. Hu, J. Xu} & 23.5770 & 24.9978 & 0.7263 & 0.7352&\ding{51} & PAM & Data
				\tabularnewline
				6 & BUAA-MC2 & \tabincell{l}{M. Xu, J. Jing, X. Deng, et al.} & 23.5733 & 24.9861 & 0.7267 & 0.7349 &\ding{51} & Optical Flow & Data
				\tabularnewline
				7 & No War & W. Guo, C. Peng, Z. Chen & 23.5664 & 24.9864 & 0.7233 & 0.7330 &\ding{51} & PAM &  Data \& Model
				\tabularnewline
				8 & GDUT\_506 & J. Chen, H. Li, J. Chen, et al. & 23.5601 & 24.9789 & 0.7239 & 0.7325 &\ding{51} & PAM & Data
				\tabularnewline
				9 & DSSR & G. Li, A. Li, L. Sun & 23.5533 & 24.9711 & 0.7242 & 0.7322 &\ding{51} & \tabincell{c}{PAM \& DConv} & Data
				\tabularnewline
				10 & xiaozhazha & D. Zhang, S. Liu & 23.5490 & 24.9570 & 0.7203 & 0.7290 &\ding{51} & \ding{55} & Data \& Model
				\tabularnewline
				11 & Zhang9678 & J. Zhang, Y. Qu & 23.5150 & 24.9346 & 0.7183 & 0.7263 &\ding{51} & PAM & \ding{55}
				\tabularnewline
				12 & NTU607QCO-SSR & H. Yang, Z. Huang, W. Chen, et al. & 23.5090 & 24.9190 & 0.7186 & 0.7265 &\ding{51}   & \ding{55}   & Data  
				\tabularnewline
				13 & supersmart & Q. Liang & 23.4896 & 24.9058 & 0.7227 & 0.7331 &\ding{51} & PAM & Data
				\tabularnewline
				14 & LIMMC\_HNU & J. Lin, Y. Wang, L. Yin, et al. & 23.4381 & 24.8550 & 0.7199 & 0.7283 &\ding{51} & PAM & \ding{55}
				\tabularnewline
				15 & HIT-IIL & R. Xu, Z. Zhang, W. Zuo & 23.4066 & 24.8165 & 0.7144 & 0.7225 &\ding{51} & \ding{55} & Data
				\tabularnewline
				16 & Hansheng & H. Guo, G. Gao, T. Zeng & 23.2918 & 24.7072 & 0.7101 & 0.7194 &\ding{51} & PAM & Model
				\tabularnewline
				17 & VIP-SSR & J. Kim, H. Kim, E. Park, J. Sim & 23.2910 & 24.7146 & 0.7103 & 0.7207 &\ding{55} & PAM & Data
				\tabularnewline
				18 & phc & H. Pi, S. Zhang & 23.2323 & 24.6584 & 0.7071 & 0.7182 &\ding{55} & PAM & \ding{55} 
				\tabularnewline
				19 & qylen & J. Zhai, P. Zeng, Y. Liu, C. Ma & 23.2241 & 24.6480 & 0.7086 & 0.7179 &\ding{51} & PAM & \ding{55}
				\tabularnewline
				20 & Modern\_SR & Y. Huang, J. Chen & 22.8370 & 24.2836 & 0.6820 & 0.6925 &\ding{55} & DConv & \ding{55}
				\tabularnewline
				\hline
				- & PASSRnet (Baseline) & - & 22.7965 & 24.2016 & 0.6801 & 0.6911 & \ding{55} & PAM & \ding{55}
				\tabularnewline
				- & Bicubic (Baseline) & - & 21.8358 & 23.3865 & 0.6287 & 0.6443 & - & - & -
				\tabularnewline
				\hline
		\end{tabular}}
	\end{table*}
	
	\section{Challenge Methods and Teams}
	\label{sec4}
	
	\subsection{The Fat, The Thin and The Young Team}

    \begin{figure}[ht]
       \centering
       \includegraphics[width=1\linewidth]{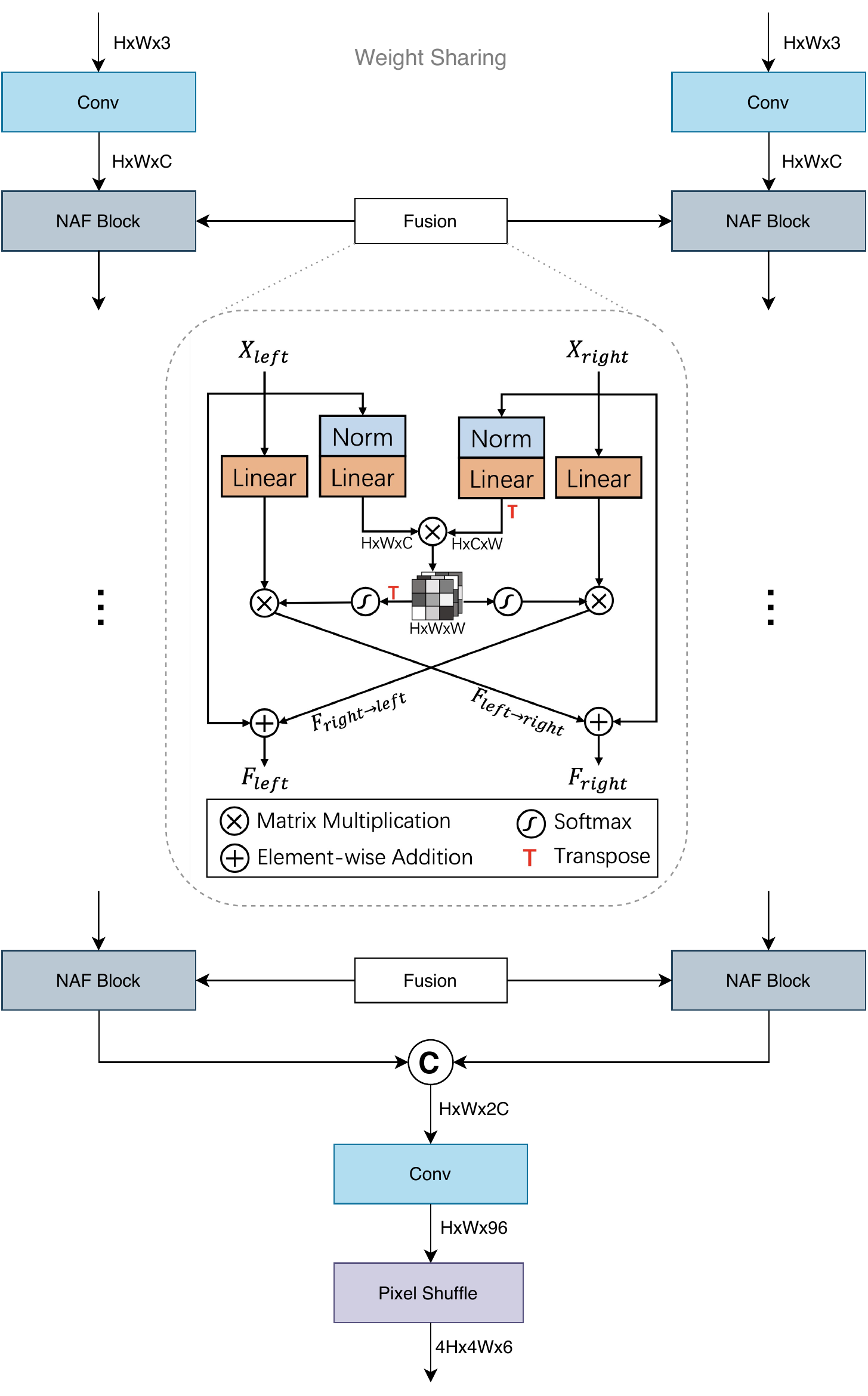}
       \caption{The Fat, The Thin and The Young Team: The network architecture of the proposed Nonlinear Activation Free Stereo image SR network (NAFSSR).} 
       \label{fig:NAFNet}
   \end{figure}
	
	This team proposed a Nonlinear Activation-Free Network (NAFNet) for image restoration \cite{chennafnet}. By using the modules in NAFNet for feature extraction, they further extended NAFNet to NAFSSR for stereo image SR, by adding cross attention modules to incorporate cross-view information. In this report, we briefly introduce their solution and readers can refer to \cite{Chu2022NAFSSR} for more details. 
	
	As shown in Fig.~\ref{fig:NAFNet}, NAFSSR has two branches with shared weights to process left and right views, respectively. Several attention modules are inserted between the left and right branches to interact cross-view information. Similar to  biPAM \cite{Wang2021Symmetric}, the attention module calculates the correlation of features along the horizontal epipolar line, and then fuses the features by performing correlation operation.
	
	In addition to the network design, a series of effective tricks were introduced to boost the SR performance. Specifically, in the training phase, random cropping, random horizontal and vertical flipping, random horizontal shifting and random RGB channel shuffling were performed for data augmentation. In the testing phase, four models were used for ensemble, and a series of test-time augmentation approaches, including horizontal and vertical flipping, RGB channel shuffling, and left-right view exchanging, were performed.
    
	Moreover, this team addressed the training/test inconsistency issue described in \cite{chu2021revisiting}, \textit{i.e.}, the training is performed on image patches while testing is performed on full image. The local-SE module in \cite{chu2021revisiting} was adopted in their solution and introduced a 0.1 dB PSNR improvement. Besides, the stochastic depth strategy \cite{huang2016deep} and the skip-init strategy \cite{de2020batch} were used to handle the over-fitting issue and facilitate the training process.
	
	\subsection{The BigoSR Team} 
\begin{figure*}[t]
        \centering
		\includegraphics[width=0.9\linewidth]{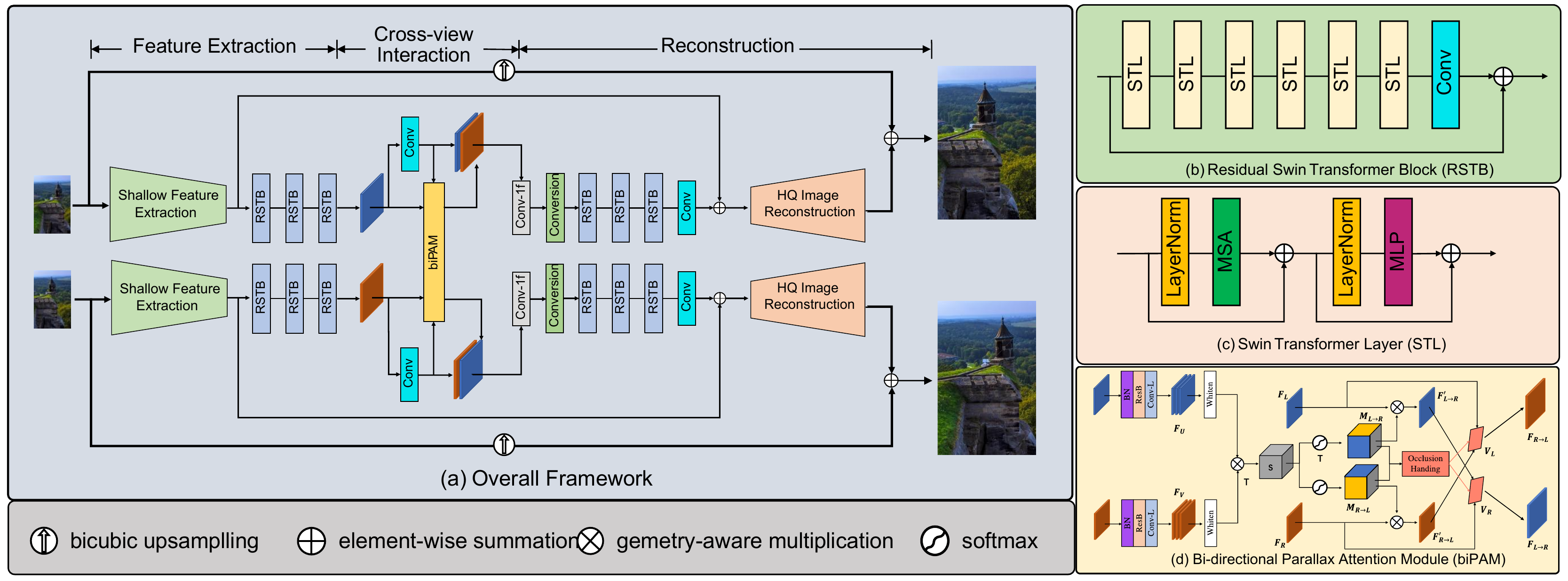}
		\caption{The BigoSR Team: The network architecture of the proposed SwiniPASSR.}
		\label{BigoSR}
	\end{figure*}

	The BigoSR team developed a SwiniPASSR network by combining the Swin Transformer \cite{Liu2021Swin} with the parallax-attention mechanism \cite{Wang2020Parallax}. To use the cross-view information from paired LR images, they employed biPAM \cite{Wang2021Symmetric} in their network. SwiniPASSR consists of three parts including feature extraction, cross-view interaction and reconstruction, as illustrated in Fig.~\ref{BigoSR}. Within the SwinIR-like framework, a biPAM module is plugged into the middle of consecutive residual swin Transformer blocks (RSTBs) to model cross-view information while handling occlussion and boundary issues. To keep semantic structure consistency with convolution-based biPAM module, a layer normalization and a patch unembedding module are used before biPAM. 
	
    
    During the training phase, to facilitate the learning of stereo correspondence, a multi-stage training strategy was employed. 
    In the first stage, stereo image pairs in the training set were divided into separate images and a Swin Transformer based network was trained for the single image SR task. At this stage, the network aims to learn structured information of images and model local spatial relationship. In the second stage, the biPAM module was plugged into middle of RSTBs to model stereo correspondence between a stereo image pair. In the third stage, the input patch size were further enlarged from 24$\times$24 to 48$\times$48 to help biPAM to aggregate cross-view information at a larger range. In the last stage, the stereo losses in the overall loss function were increased by 10 times for fine-tuning to encourage the network to focus more on cross-view information.
	
	\subsection{The NUDT-CV\&CPLab Team}
	
   \begin{figure*}[htbp]
   \centering
   \includegraphics[width=0.9\linewidth]{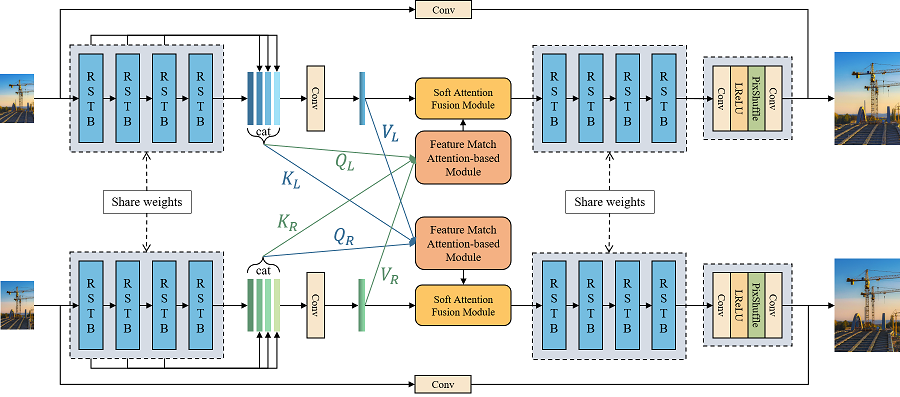}
   \caption{The NUDT-CV\&CPLab Team: The network architecture of the proposed SSRFormer.} 
   \label{NUDT}
   \end{figure*}
   

   Inspired by SwinIR, the NUDT-CV\&CPLab team proposed a Transformer-based network architecture (namely, SSRFormer) for stereo image SR, as shown in Fig.~\ref{NUDT}. SSRFormer is a Siamese network architecture with two branches sharing weights.  Specifically, four residual Swin Transformer blocks (RSTBs) are first used as the feature extractor to extract deep features. Then, inspired by parallax attention mechanism, an attention-based feature matching (AFM) module is adopted to extract rich cross-view information without explicitly aligning the left and right images. 

	During the training phase, 800 pairs of stereo images were used as the training set. HR images were randomly cropped into $192 \times 192$ patches, and LR images were cropped accordingly. Random flipping was used for data augmentation. The proposed SSRFormer was first trained for 300,000 iterations on two 2080ti GPUs with batch size of 8 using the L1 loss. Then, the model was further fine-tuned for 124,000 iterations on four 2080ti GPUs with batch size of 16. An L1 loss was adopted for the first 60000 iterations and an L2 loss was used for the remaining iterations. The learning rate was initialized to $2\times10^{-4}$ and halved at iteration 250000, 300000, 375000, and 400000.

   \subsection{The BUPT-PRIV Team}
   \begin{figure}
   \centering
   \includegraphics[width=1\linewidth]{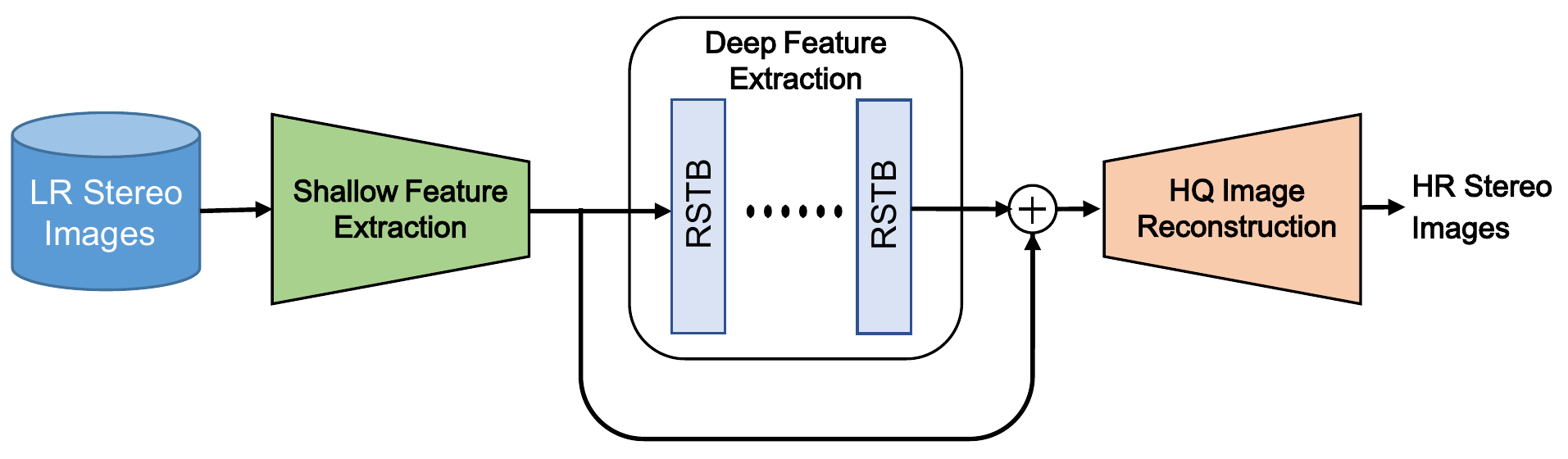}
   \caption{The BUPT-PRIV Team: The network architecture of the proposed SwinIR-impr network.} 
   \label{fig:caopulan}
   \end{figure}
	
	The BUPT-PRIV team developed an improved version of SwinIR \cite{liang2021swinir} to super-resolve left and right images, respectively. The network architecture is shown in Fig.~\ref{fig:caopulan}. Although the cross-view information is not used in this solution, benefited from the effective data augmentation and test-time augmentation strategies, the proposed method achieves a very competitive SR performance. In addition to the data augmentations originally used in SwinIR, this team further introduced a series of tricks. \textit{In the training phase}, they: 1) adjusted the possibility of selecting training samples to ensure that an image with higher resolution will get a higher possibility to be selected, 2) randomly shuffled RGB channels with a probability of 50\%, and 3) trained three models with different combinations of architectures and losses. \textit{In the testing phase}, a series of test-time augmentation approaches was adopted including flipping, self-ensemble, and RGB shuffling. Note that, the window size was set to 16 in this method, which is different from the setting (\textit{i.e.}, 8) in SwinIR.

	\subsection{The NKU\_caroline Team}
	
\begin{figure}[t]
        \includegraphics[width=1\linewidth]{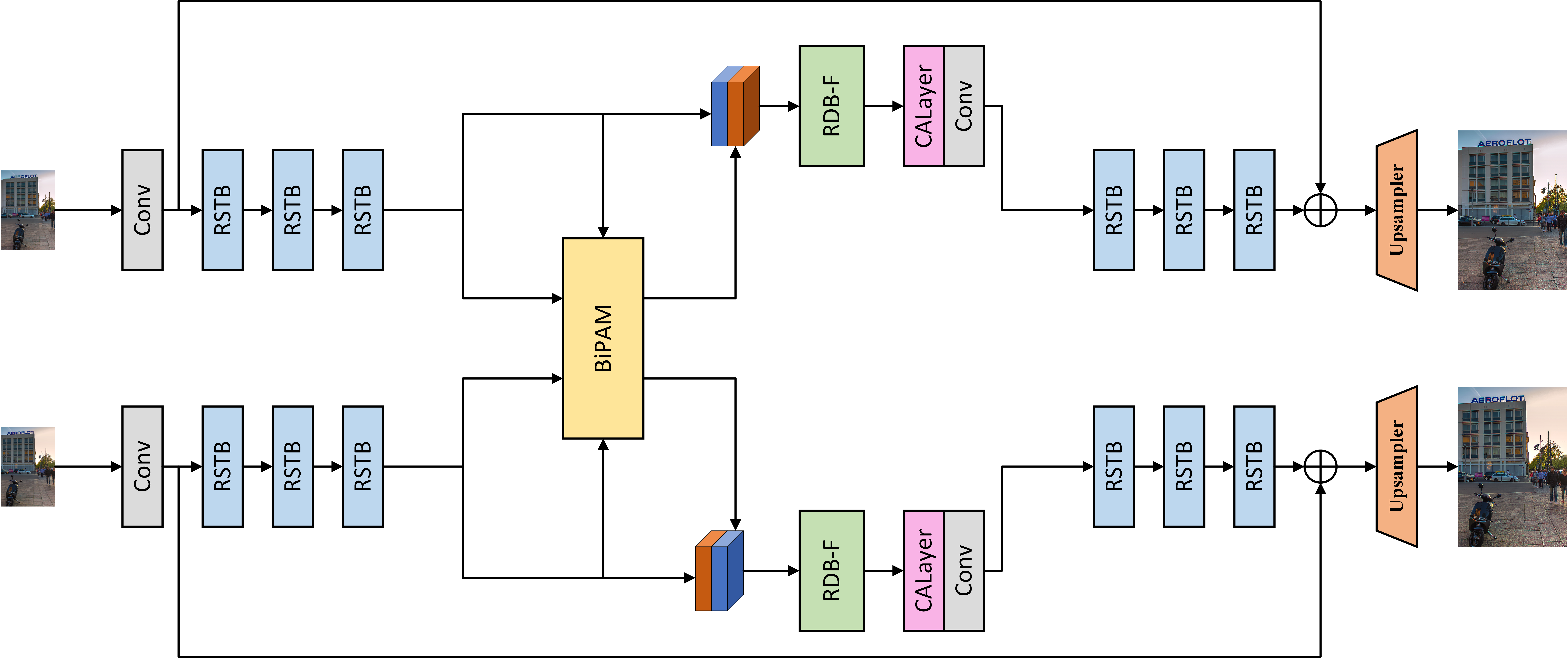}
        \caption{The NKU\_caroline Team: The network architecture of the proposed PAMSwin network.}
        \label{NKU_caroline}
    \end{figure}

	The NKU\_caroline team shares a similar idea with many other teams and developed a PAMSwin network by combining SwinIR \cite{liang2021swinir} with parallax-attention mechanism \cite{Wang2019Learning} for the stereo image SR task. The network architecture of PAMSwin is shown in Fig.~\ref{NKU_caroline}. Within the SwinIR framework, a biPAM module is plugged into the middle of residual swin Transformer blocks (RSTBs) to capture cross-view information. Besides, a channel attention layer is employed to exploit correlations between different channels. This team also emphasizes that the order of the input left and right images contains priori information and is critical to the performance. {Training with mixed orders of left-right images produced inferior SR performance in their experiments.}
	
	
	During the training phase, a three-stage training strategy was employed. First, the proposed PAMSwin was trained from scratch for 500K iterations. Then, Cutblur~\cite{yoo2020rethinking} was included for data augmentation to fine-tune the model with the best performance in the first stage for 500K iterations. Note that, the parameters for the biPAM module were fixed at this stage. Finally, a small learning rate was used to further fine-tune the whole model with the highest SR accuracy in the second stage for 500K iterations. During the testing phase, a self-ensemble strategy was adopted to improve the performance.

	\subsection{The BUAA-MC2 Team}
	
   \begin{figure}[htbp]
   \centering
   \includegraphics[width=1\linewidth]{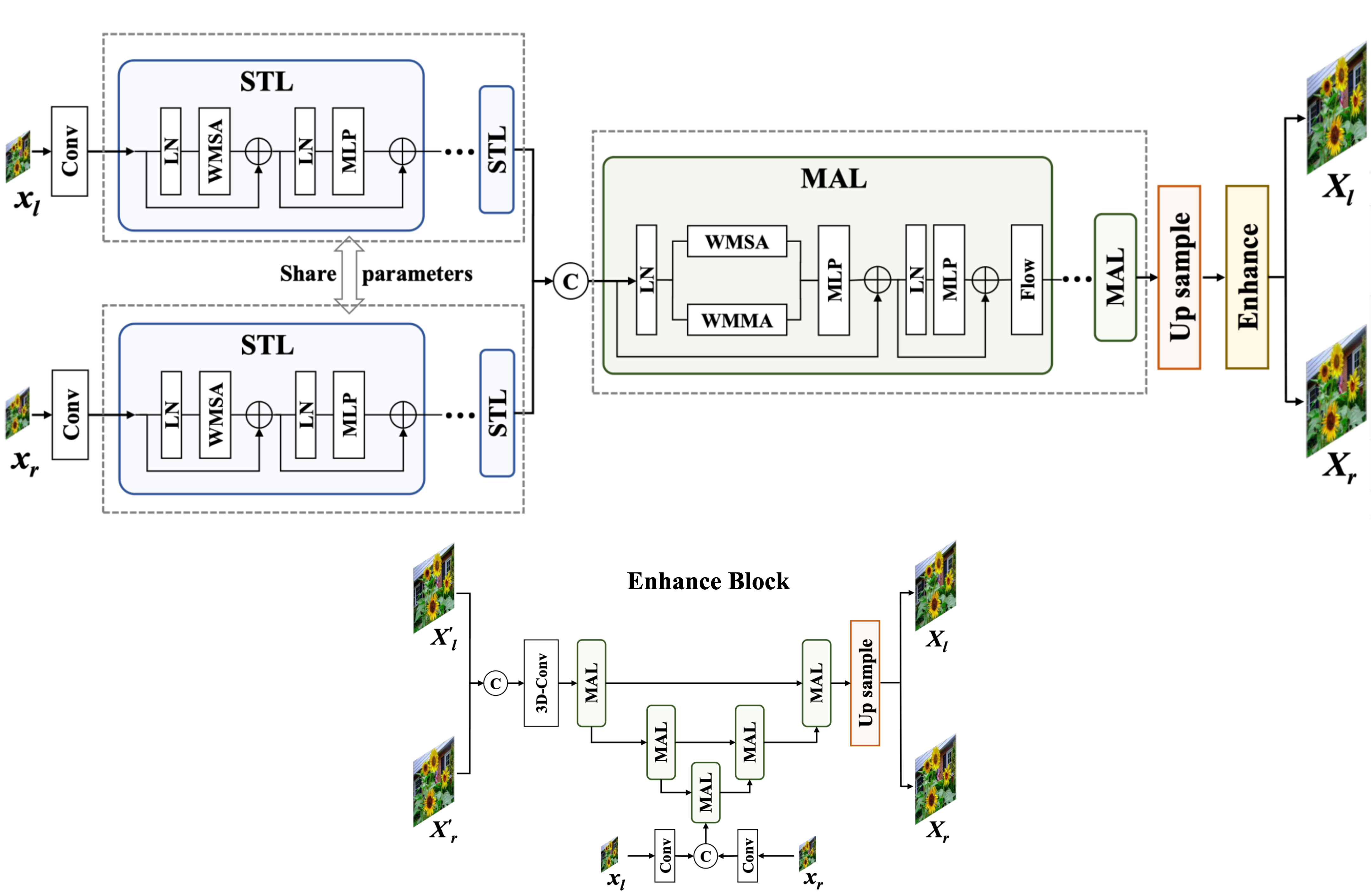}
   \caption{The BUAA-MC2 Team: The architecture of the proposed Stereo Image Super-Resolution Transformer (StereoSRT).} 
   \label{BUAA-StereoSRT}
   \end{figure}
   
	The BUAA-MC2 team proposed a Stereo Image Super-Resolution Transformer (StereoSRT). As shown in Fig.~\ref{BUAA-StereoSRT}, the input stereo images are first fed into a shallow convolution layer and several Swin Transformer layers (STL) to extract shallow features. Then, the output feature maps are concatenated and passed to several mutual attention layers (MAL) to extract cross-view information. After MALs, HR images are reconstructed using a sub-pixel convolutional layer. Finally, a multi-scale enhancement module consisting of several MALs is adopted to further enhance the quality of the HR images. 
	
	During the training phase, an L1 loss was used for SR and an L2 loss was used for enhancement. The initial learning rate was set to $4\times10^{-4}$. The model was trained with a multi-stage strategy. Specifically, in the first stage, the model was trained only with the STL part (the output of STL wass directly fed into the up-sample module) for 200K iterations with a patch size of 64$\times$64. In the second stage, the MAL part (without the flow module) was optimized for 200K iterations while the parameters of the STL parts were fixed. In the third stage, the whole network was optimized end-to-end for 100K iterations. In the fourth stage, the flow module was added to MAL and optimized for 300K iterations with the STL parts being fixed. Finally, the whole model was fine-tuned with a patch size of 96 $\times$ 96 for 100K iterations.

	\subsection{The No War Team}
	\begin{figure}
    \centering
    \includegraphics[width=1\linewidth]{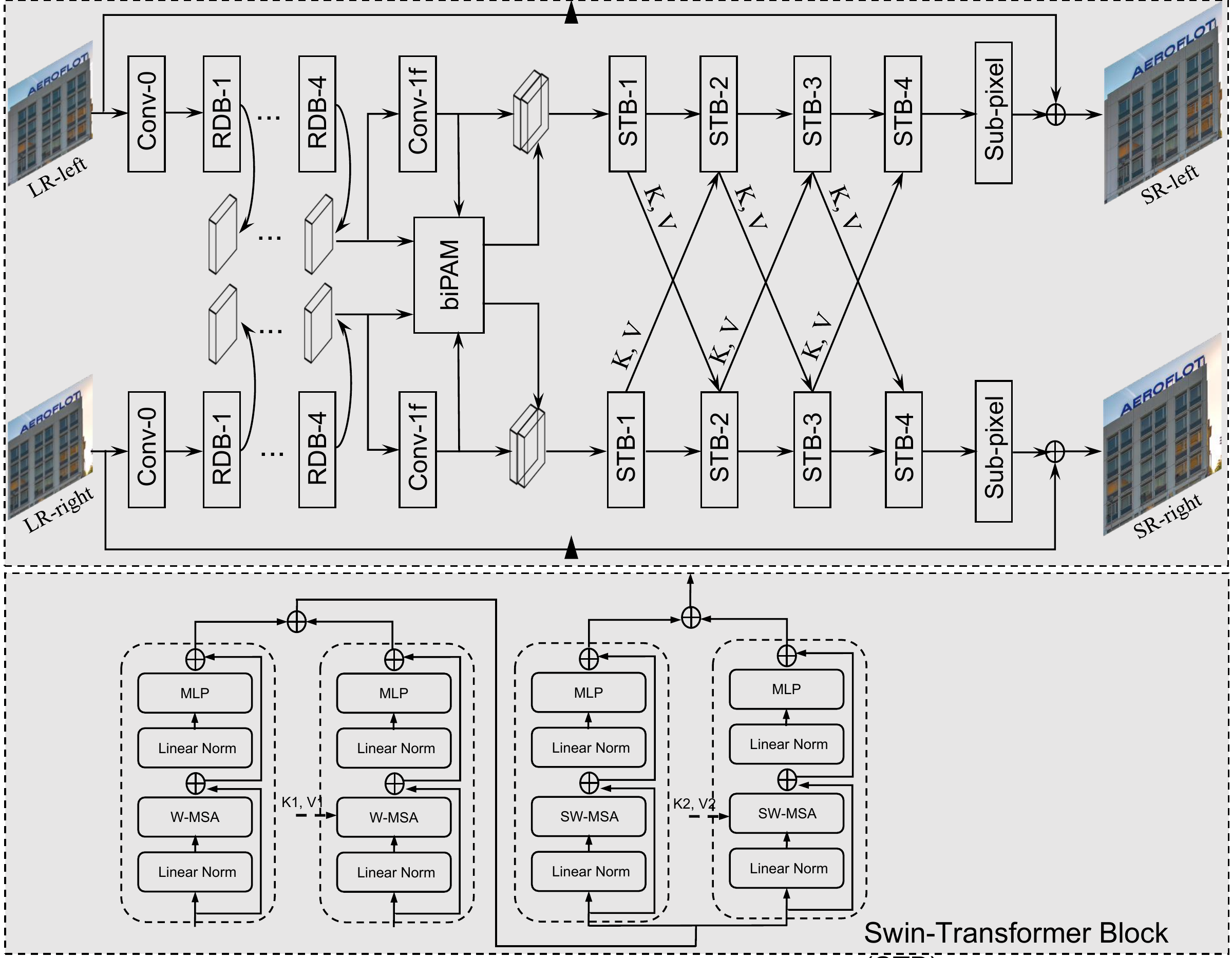}
   \caption{The No War Team: The architecture of the proposed Parallel Interactive Transformer (PAIT).} 
   \label{fig:PAIT}
   \end{figure}
	
	Figure~\ref{fig:PAIT} illustrates the network architecture proposed by the NO War team. The participants of this team used the modules of iPASSR \cite{Wang2021Symmetric} for feature extraction and cross-view information interaction. In the reconstruction stage, a parallel interactive Transformer with four Swin Transformer Blocks (STBs) \cite{Liu2021Swin} was designed to further enhance the interaction of left and right view features. During the training phase, the input images were cropped into patches of size $128\times128$ with a stride of 20. The batch size was set to 4. To prevent model over-fitting, the participants performed model ensemble by selecting five models at non-adjacent epochs and averaging their weights to obtain the final model for test. 

	\subsection{The GDUT\_506 Team}
	
\begin{figure}[t]
        \centering
        \includegraphics[width=1\linewidth]{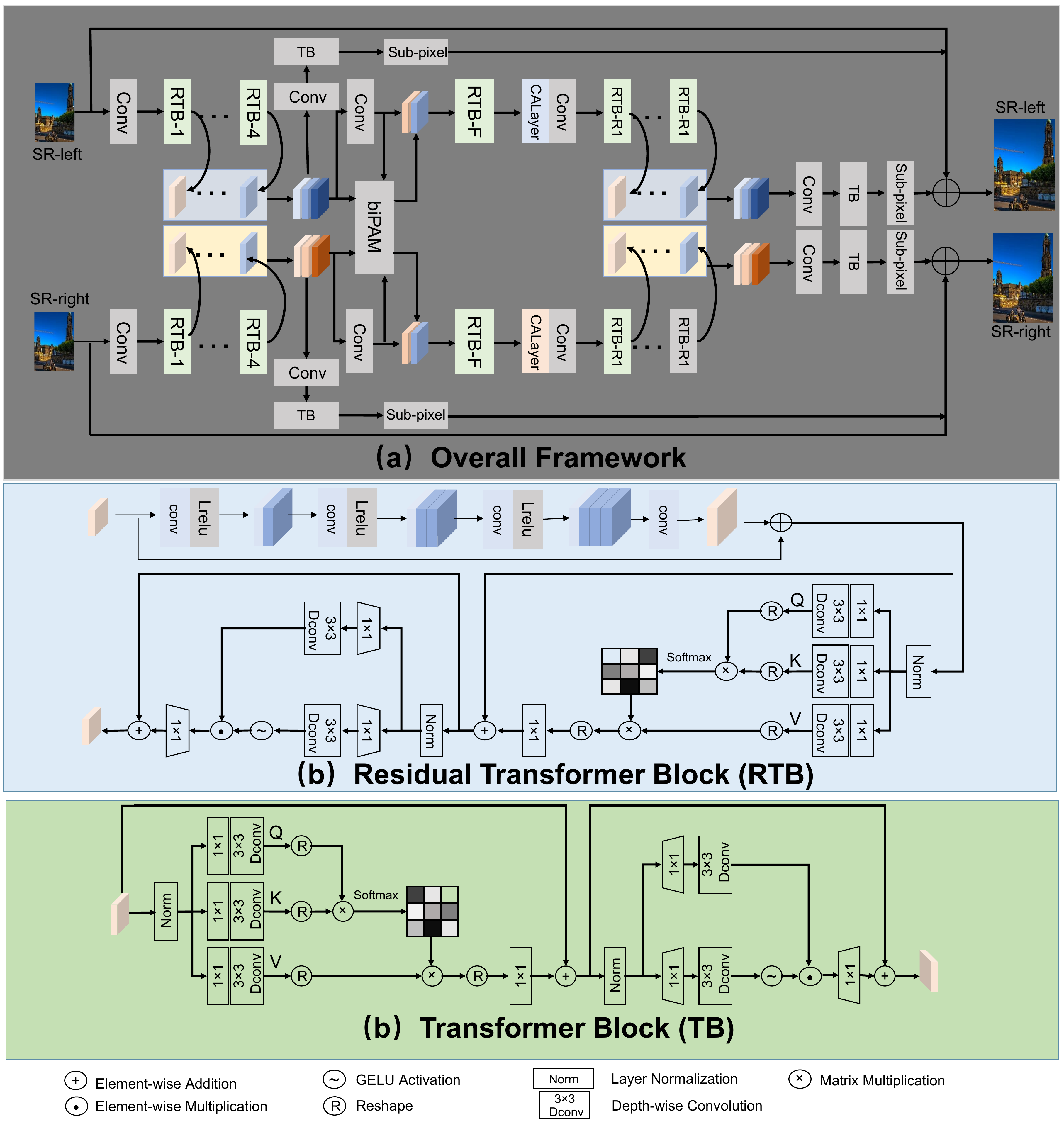}
        \caption{The GDUT\_506 Team: The network architecture of the proposed PRTN.}
        \label{GDU_506}
    \end{figure}

	The GDUT\_506 team developed a Parallax Res-Transformer Network (PRTN) by combining Transformers with the parallax attention mechanism. The network architecture of the proposed PRTN is illustrated in Fig.~\ref{GDU_506}. First, four residual Transformer blocks (RTBs) are used for shallow feature extraction. Then, a biPAM module is employed to perform cross-view interaction between features extracted from left and right images. Finally, four cascaded RTBs are used to obtain deep features and a Transformer Block (TB) is adopted to reconstruct the HR images. Note that, the proposed RTB is hybrid module mixed with both convolutional layers and Transformer layers.
	
	
    During the training phase, random horizontal and vertical flipping was used for data augmentation. A three-stage training strategy was employed. At the first stage, PRTN was trained for $\times$2 SR using an L1 loss. At the second stage, PRTN was fine-tuned for $\times$ 4 SR using an L1 loss. At the last stage, PRTN was further fine-tuned for $\times$ 4 SR using an L2 loss.
    
    
    
    
    

	\subsection{The DSSR Team}
	
	\begin{figure}[t]
        \includegraphics[width=1\linewidth]{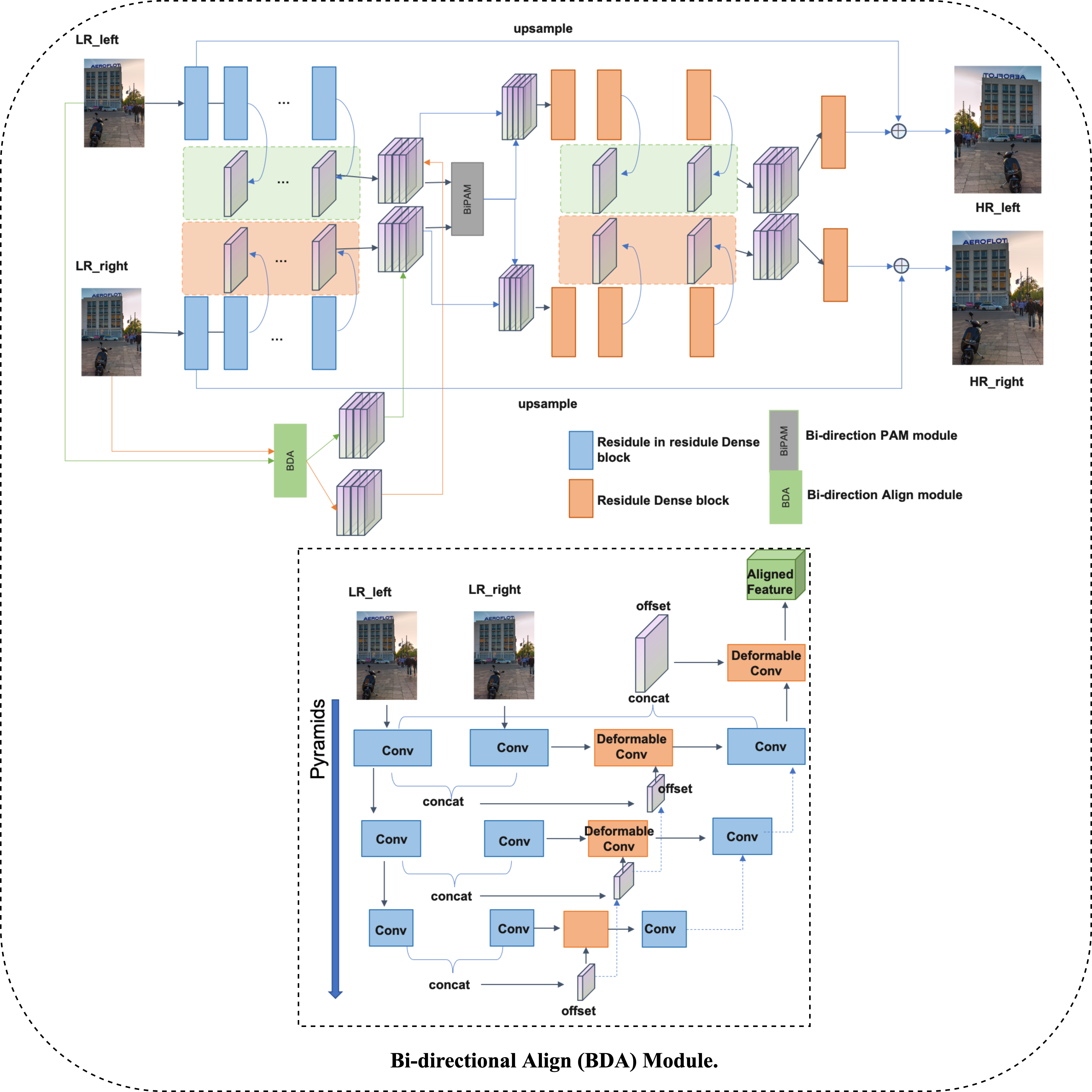}
        \caption{The DSSR Team: The architecture of the proposed Deformable Stereo Super-Resolution (DSSR).}
        \label{DSSR}
    \end{figure}
    
    The DSSR team proposed a Deformable Stereo Super-Resolution (DSSR) network. As shown in Fig.~\ref{DSSR}, DSSR consists of a coarse SR stage and a refinement stage. In the coarse SR stage, a Bi-Directional Align (BDA) module with pyramid cascading deformable convolutions \cite{Wang2019EDVR} is used to help the biPAM module \cite{Wang2021Symmetric} to better exploit cross-view information. 
    In the refinement stage, the SR results from the previous stage is fed into a refinement module with 4 groups of residual in residual dense block (RRDB) \cite{Wang2018ESRGAN} for enhancement.
    
    During the training phase, the generated LR images were cropped into patches of size $120 \times 120$ with a stride of 40. These patches were randomly flipped horizontally and vertically for data augmentation. All models were optimized using the Adam method with $\beta_1 = 0.9$, $\beta_2 = 0.999$, and a batch size of 36. The initial learning rate was set to $2\times10^{-4}$ and reduced to half after every 30 epochs. The training was stopped after 100 epochs. In the early stage of training,  an L1 loss was used to accelerate convergence. Then, an L2 loss was used to obtain higher results in terms of PSNR.

	\subsection{The Xiaozhazha Team}
	
	\begin{figure}[t]
		\centering
		\includegraphics[width=1\linewidth]{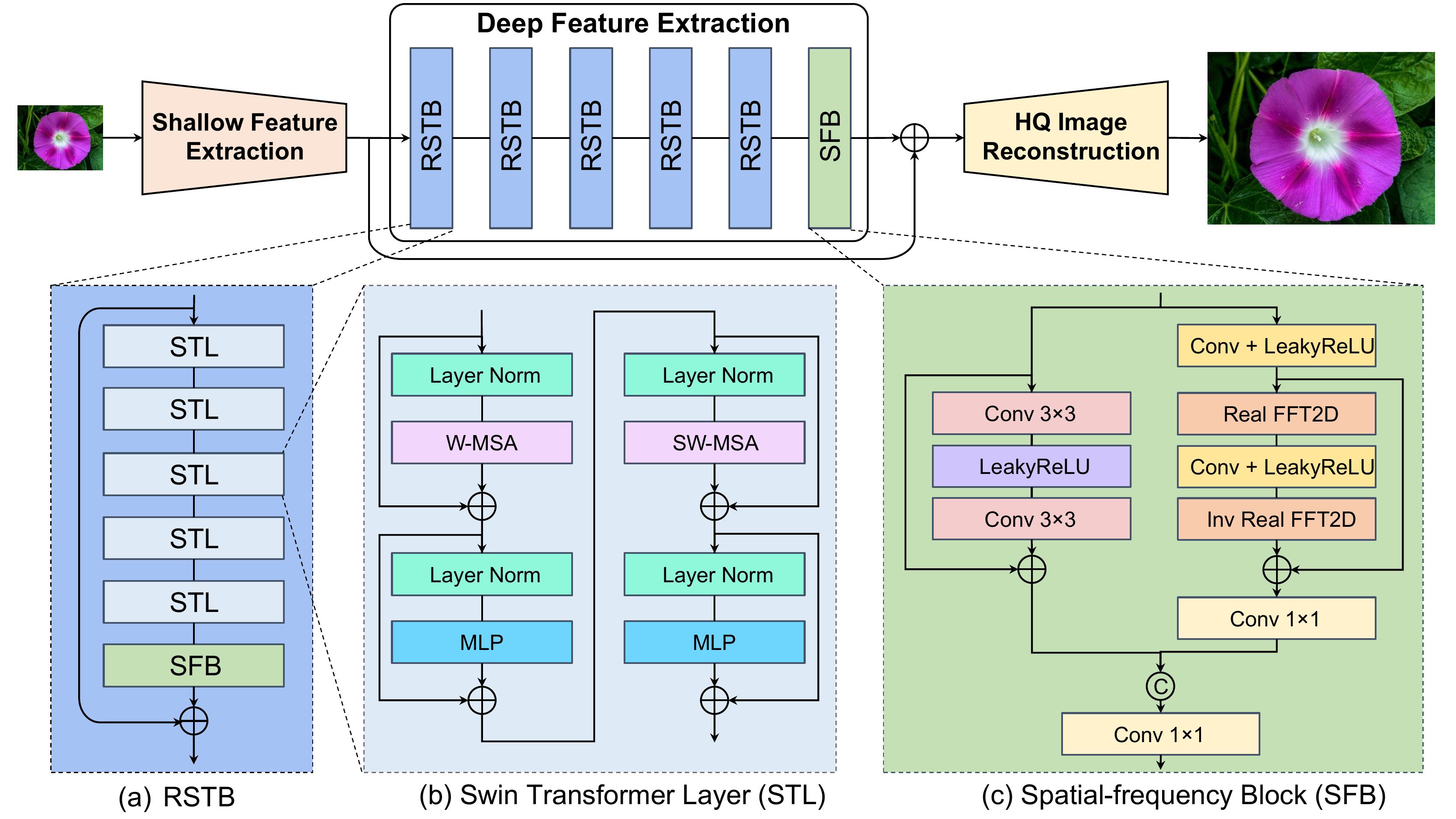}
		\caption{The Xiaozhazha Team: The network architecture of the proposed SwinFIR for stereo image super-resolution.}
		\label{fig:xiaozhazha}	
	\end{figure}
	
	The xiaozhazha team proposed a network called SwinFIR based on SwinIR \cite{liang2021swinir} and fast Fourier convolution~\cite{chi2020fast}, as shown in Fig.~\ref{fig:xiaozhazha}. SwinFIR consists of three modules, including a shallow feature extraction module, a deep feature extraction module and a high-quality image reconstruction module. The shallow feature extraction and high-quality image reconstruction modules adopt the same configurations as in SwinIR \cite{liang2021swinir}. Since the fast Fourier convolution can extract global features, the participants replaced the 3$\times$3 convolution in SwinIR with fast Fourier convolution and a residual module to fuse global and local features. The proposed spatial-frequency block improves the representation capability of this model. 
	
	During the training phase, random horizontal flipping, random vertical flipping,  random RGB channel shuffling and mix-up strategy \cite{yoo2020rethinking} were used for data augmentation. Self-ensemble and multi-model ensemble were adopted to further improve the SR performance.

	\subsection{The Zhang9678 Team}
	
\begin{figure}[t]
		\centering
		\includegraphics[width=1\linewidth]{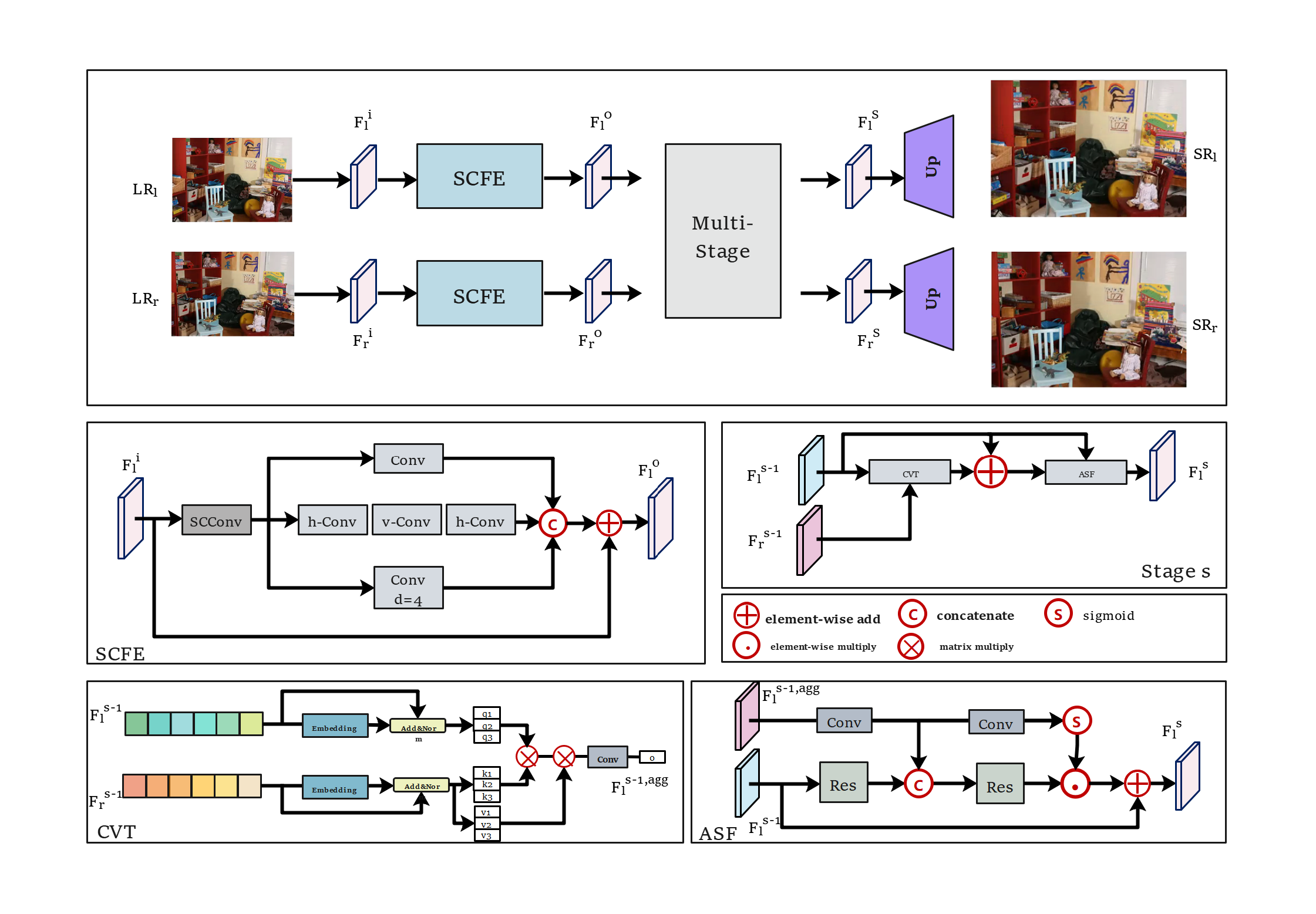}
		\caption{The Zhang9678 Team: The network architecture of the proposed MPTnet.}
		\label{Zhang9678}	
	\end{figure}

	The Zhang9678 team developed a multi-stage progressive Transformer network (MPTnet) for stereo image SR. The network architecture of the proposed MPTnet is shown in Fig.~\ref{Zhang9678}. First, self-calibrated feature extractor (SCFE) is used for feature extraction. Within each SCFE, SCConv \cite{Liu2020Improving} and a three-branch structure are employed to achieve large receptive fields. Then, multiple cross-view Transformers (CVTs) and adaptive selective modules (ASFs) are employed to exploit cross-view information. CVT performs information interaction between left and right images along epipolar lines, while ASF aggregates features from different views using a gating mechanism.
	
    
	\subsection{The NTU607QCO-SSR Team}
	
\begin{figure}[t]
		\centering
		\includegraphics[width=1\linewidth]{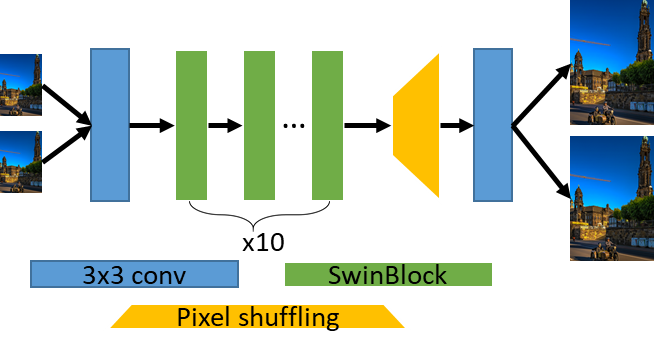}
		\caption{The NTU607QCO-SSR Team: The network architecture of the proposed model.}
		\label{NTU607QCO}	
	\end{figure}

	The NTU607QCO-SSR team mainly considers the stereo image super-resolution task as a single image super-resolution task and adopts the state-of-the-art SwinIR~\cite{liang2021swinir} as the backbone. As shown in Fig.~\ref{NTU607QCO}, the model contains convolutional blocks, SwinBlocks, and a pixel shuffling layer. Images are first passed to the $3\times3$ convolutions and then SwinBlocks are used to extract the global and local features. At the end of the SwinBlocks, image features are passed to a pixel shuffling layer and a $3\times3$ convolution is used to enlarge the feature maps and reconstruct the SR result.
	During the training phase, an L1 loss was first used for optimization with 300 epochs. After that, a wavelet-based L1 loss was adopted for fine-tuning. The wavelet-based loss~\cite{yang2020net,chen2021contourletnet} uses wavelet transforms to generate sub-images with different scales and frequencies from the original image. Since the resultant sub-images have higher-frequency details, better performance can be achieved.
	
	\subsection{The Supersmart Team} 
	
	\begin{figure}[t]
		\centering
		\includegraphics[width=1\linewidth]{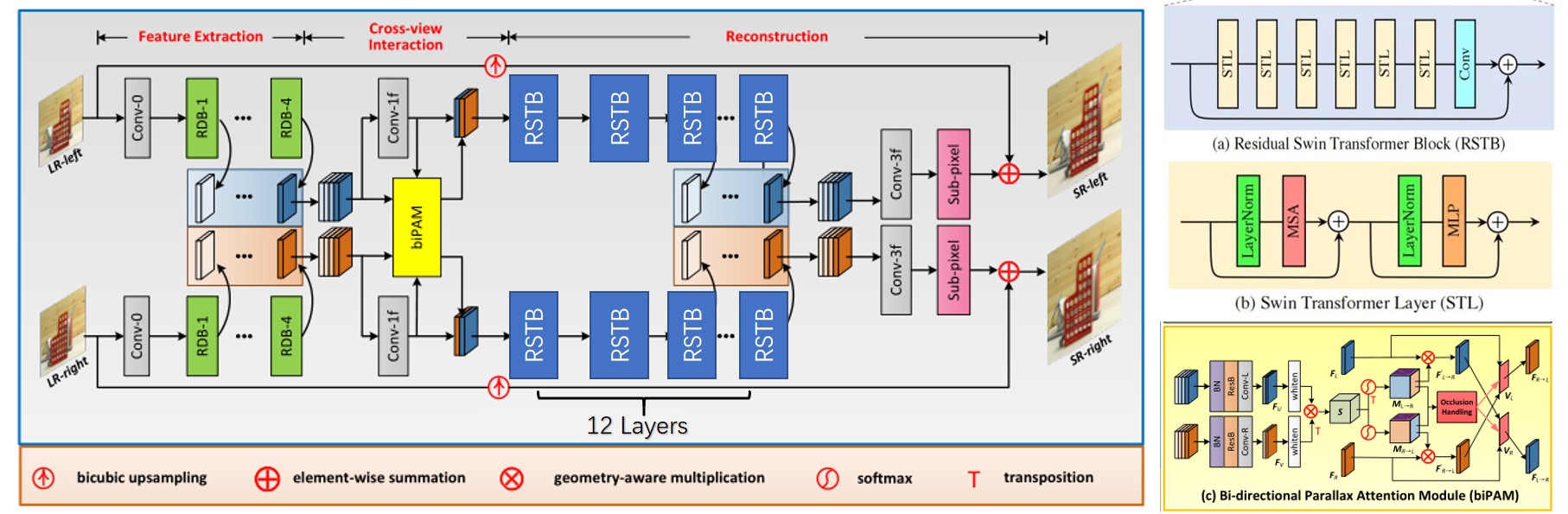}
		\caption{The supersmart Team: The network architecture of the proposed SwinRSTB.}
		\label{fig:SwinRSTB}	
	\end{figure}
	
	The supersmart team proposed a method called SwinRSTB, as shown in Fig.~\ref{fig:SwinRSTB}. Since SwinIR \cite{liang2021swinir} is designed for single image SR and cannot incorporate cross-view information, this team combined iPASSR \cite{Wang2021Symmetric} with SwinIR for stereo image SR. In the proposed SwinRSTB network, the RSTB module in SwinIR was used to replace the RGB module in iPASSR. 
	
	\subsection{The LIMMC\_HNU Team}
	
\begin{figure}[t]
        \centering 
        \includegraphics[width=1\linewidth]{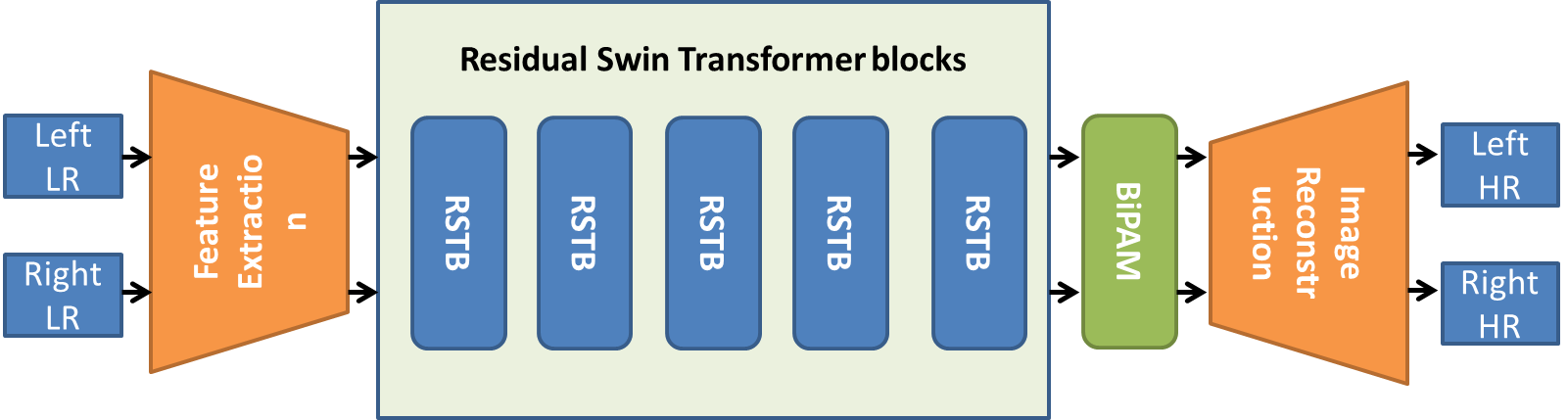}
        \caption{The LIMMC\_HNU Team: The network architecture of the proposed PAMSwinIR.}
        \label{LIMMC_HNU}
    \end{figure}

	The LIMMC\_HNU team developed a PAMSwinIR network inspired by SwinIR \cite{liang2021swinir} and iPASSRnet \cite{Wang2021Symmetric}. The network architecture is illustrated in Fig.~\ref{LIMMC_HNU}. Different from the solutions of many other teams, they postponed the biPAM module until the end of the residual swin Transformer blocks. During the training phase, the loss function in \cite{Wang2021Symmetric} was first used for training to capture cross-view correspondence. Then, only MSE loss was adopted for fine-tuning.
	

	\subsection{The HIT-IIL Team}
	
   \begin{figure}
   \centering
   \includegraphics[width=1\linewidth]{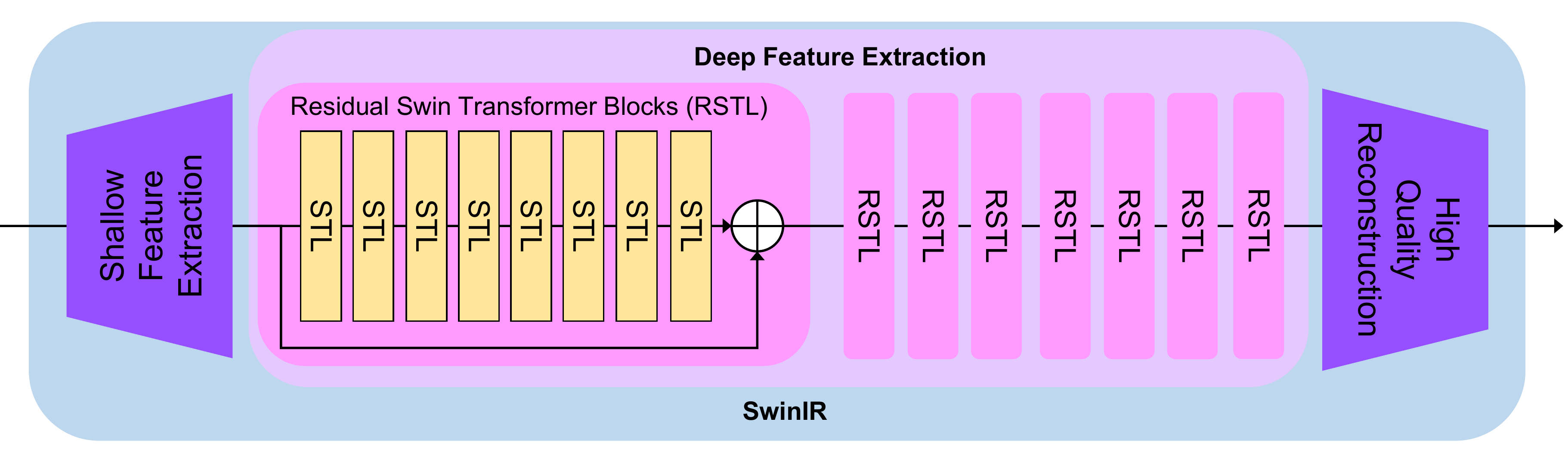}
   \caption{The HIT-IIL Team: Network architecture of SwinIR.} 
   \label{HITIIL}
   \end{figure}
   
   The HIT-IIL team employed SwinIR~\cite{liang2021swinir} (see Fig.~\ref{HITIIL}) as a basic SISR model and introduced an FFT loss for optimization. The FFT loss measures the difference between outputs and their corresponding HR image in the frequency domain:
	\begin{equation}
	    \centering
	    \mathcal{L}_\mathrm{FFT}(\mathbf{y}, \hat{\mathbf{y}})  =  \| \mathbf{F}(\mathbf{y}) - \mathbf{F}(\hat{\mathbf{y}}) \|_1,
	\end{equation}
	where $\mathbf{F}$ denotes the Fourier transform, $\mathbf{y}$ is the output of the model, and $\hat{\mathbf{y}}$ represents the ground truth image. Therefore, the overall loss function can be written as:
	\begin{equation}
	    \centering
	    \mathcal{L}_\mathrm{total}(\mathbf{y}, \hat{\mathbf{y}})  = \| \mathbf{y} - \hat{\mathbf{y}} \|_1 + \lambda * L_{FFT}(\mathbf{y}, \hat{\mathbf{y}}) .
	\end{equation}
	Compared with the model trained with only L1 loss, the additional FFT loss helps the model converge faster and obtain higher performance.
	
	\subsection{The Hansheng Team}
	\begin{figure}[t]
		\centering
		\includegraphics[width=1\linewidth]{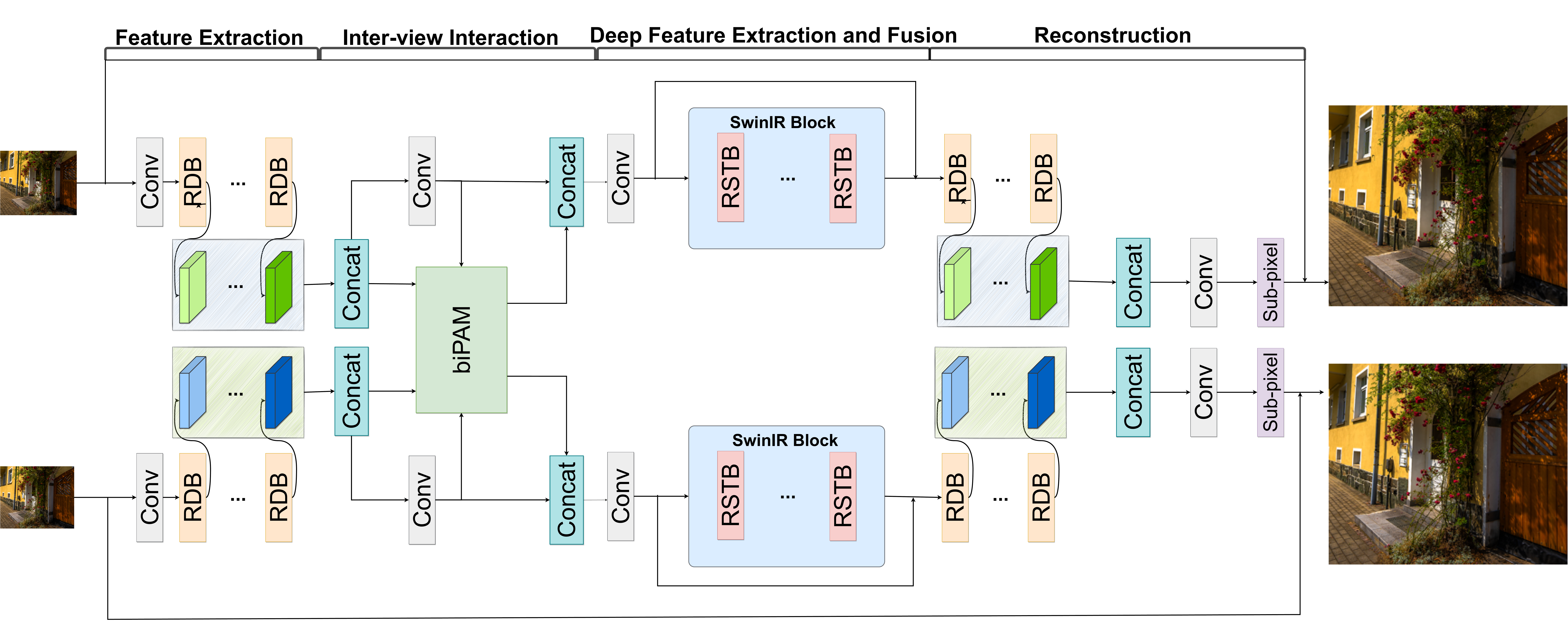}
		\caption{The Hansheng Team: The network architecture of the proposed fine-tuned SwinIR.}
		\label{fig:HanSheng}
	\end{figure}
	
	The Hansheng team developed a stereo image SR network based on SwinIR \cite{liang2021swinir} and iPASSR \cite{Wang2021Symmetric}. As shown in Fig.~\ref{fig:HanSheng}, the proposed network first performs feature extraction and cross-view interaction using the modules of iPASSR. Then, 6 RSTB blocks are used to aggregate the left and right features, with each block consisting of 6 STL blocks. Next, the resultant features are further fed into several RDBs for reconstruction. During the training phase, input images were cropped into patches of 48$\times$48, and the window size in STL blocks was set to 8.
	
	\subsection{The VIP-SSR Team}
	
\begin{figure}[t]
		\centering
		\includegraphics[width=1\linewidth]{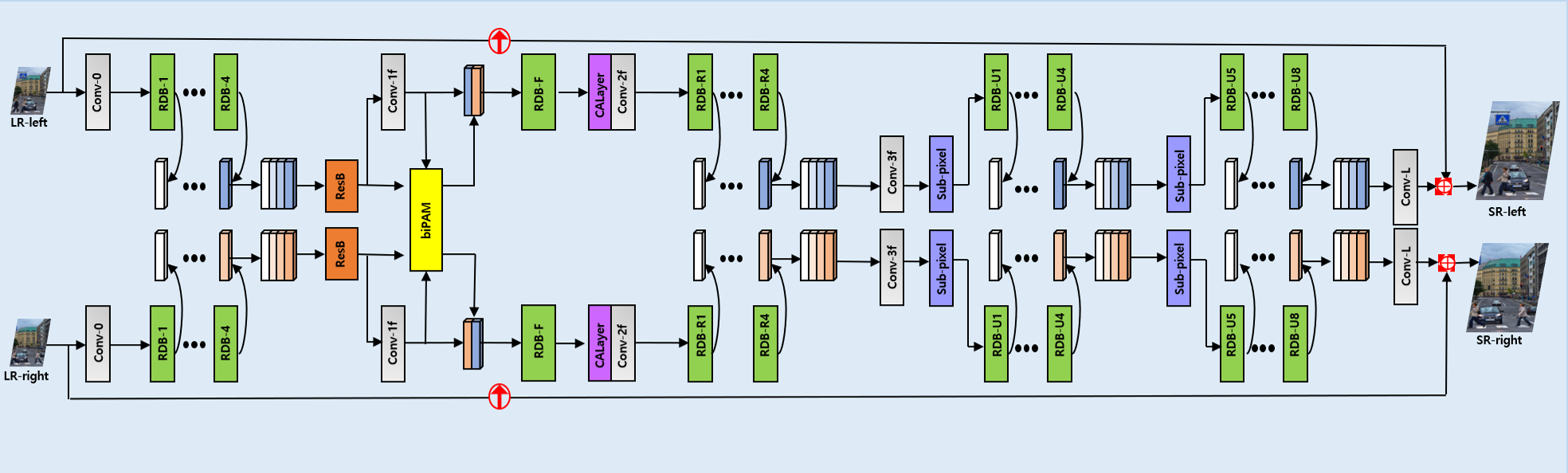}
		\caption{The VIP-SSR Team: The network architecture of the proposed HFB-iPASSR.}
		\label{VIP-SSR}
	\end{figure}

	The VIP-SSR team improved the performance of iPASSRnet \cite{Wang2021Symmetric} by introducing a hierarchical feature blended-iPASSR (HFB-iPASSR) network. They first split the $\time4$ pixel-shuffle layer in iPASSR into two $\time2$ pixel-shuffle layers, with each one being followed by a residual dense group (RDG) to relax the discontinuity along pixels caused by the pixel-shuffle operation. They also add a residual block to the first RDG as post-processing to utilize the relation of multi-level features. The architecture of the proposed HFB-iPASSR network is shown in Fig.~\ref{VIP-SSR}.
	
    \subsection{The phc Team}
	
   \begin{figure}[t]
   \centering
   \includegraphics[width=1\linewidth]{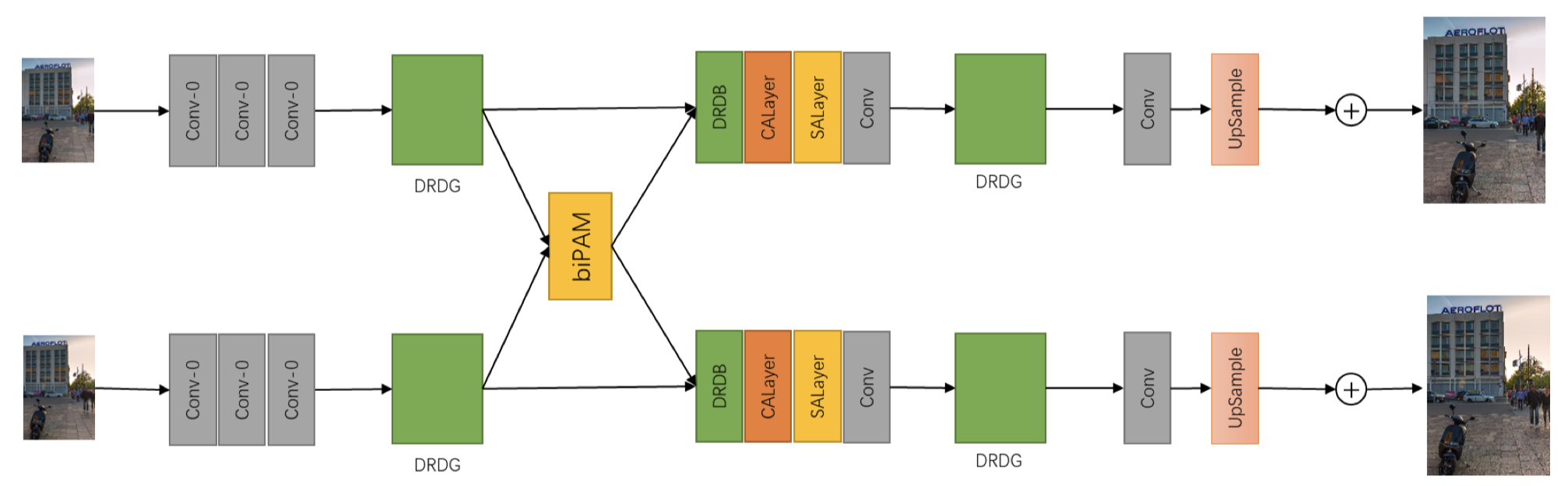}
   \caption{The phc Team: Network architecture of Improved-PASSR.} 
   \label{phc}
   \end{figure}
   
   Inspired by PASSRnet~\cite{Wang2019Learning,Wang2020Parallax}, the phc team proposed an Improved-PASSR, as shown in Fig.~\ref{phc}. Specifically, the participants introduced a self-attention module to further capture long-range correlations within the image. Meanwhile, the participants deepen the original RDB layers and employ a pixel perception block (PPB) for feature enhancement. In the upsampling block, two sub-pixel layers are used to generate the super-resolved image gradually. Since batch normalization cannot introduce a notable performance improvement, it is removed from the network. During the training phase, the model was optimized for 100 epochs using the Adam method with $\beta_1 = 0.9$, $\beta_2 = 0.999$ and a batch size of 32. The initial learning rate was set to $2\times10^{-4}$ and reduced to half after every 40 epochs.
	
	\subsection{The qylen Team}
	
	\begin{figure}[h]
		\centering
		\includegraphics[width=1\linewidth]{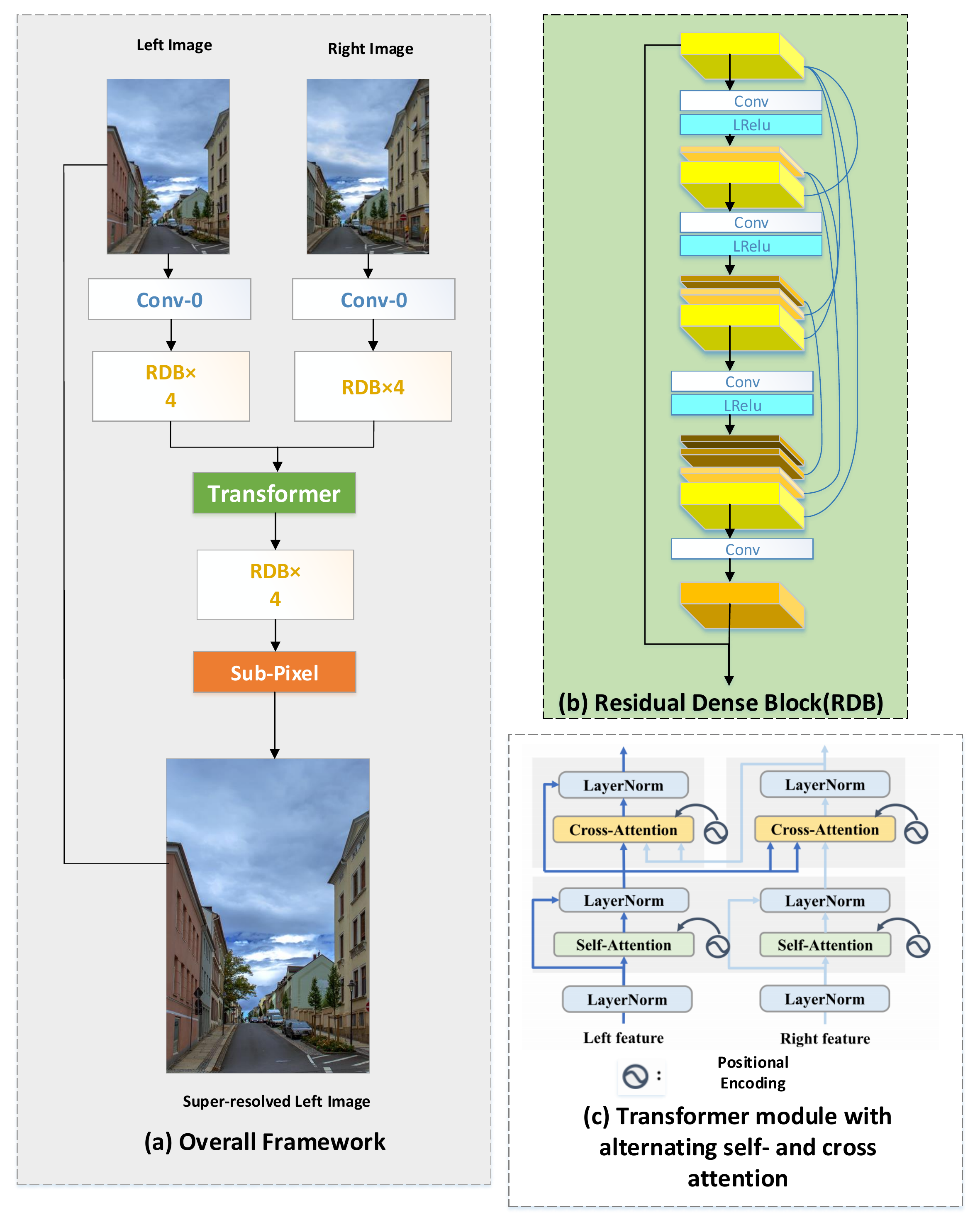}
		\caption{The qylen Team: The network architecture of the proposed iPASSR-Transformer Net.}
		\label{fig:qylen}
	\end{figure}
	
	The qylen team combined iPASSR with Transformers to achieve improved SR performance. As shown in Fig.~\ref{fig:qylen}, the proposed network sequentially performs feature extraction, Transformer-based information fusion, and SR reconstruction. The feature extraction and SR reconstruction parts of this method are similar to those in iPASSR. In the proposed Transformer-based information fusion module, self-attention is used to model dependencies among pixels within a single view, while cross attention is used to model correspondence of pixels along the epipolar lines between two views. These two attention modules are alternately applied to update the features based on intra-view and cross-view information.
	
	\begin{figure}[t]
	    \centering
	    \includegraphics[width=1\linewidth]{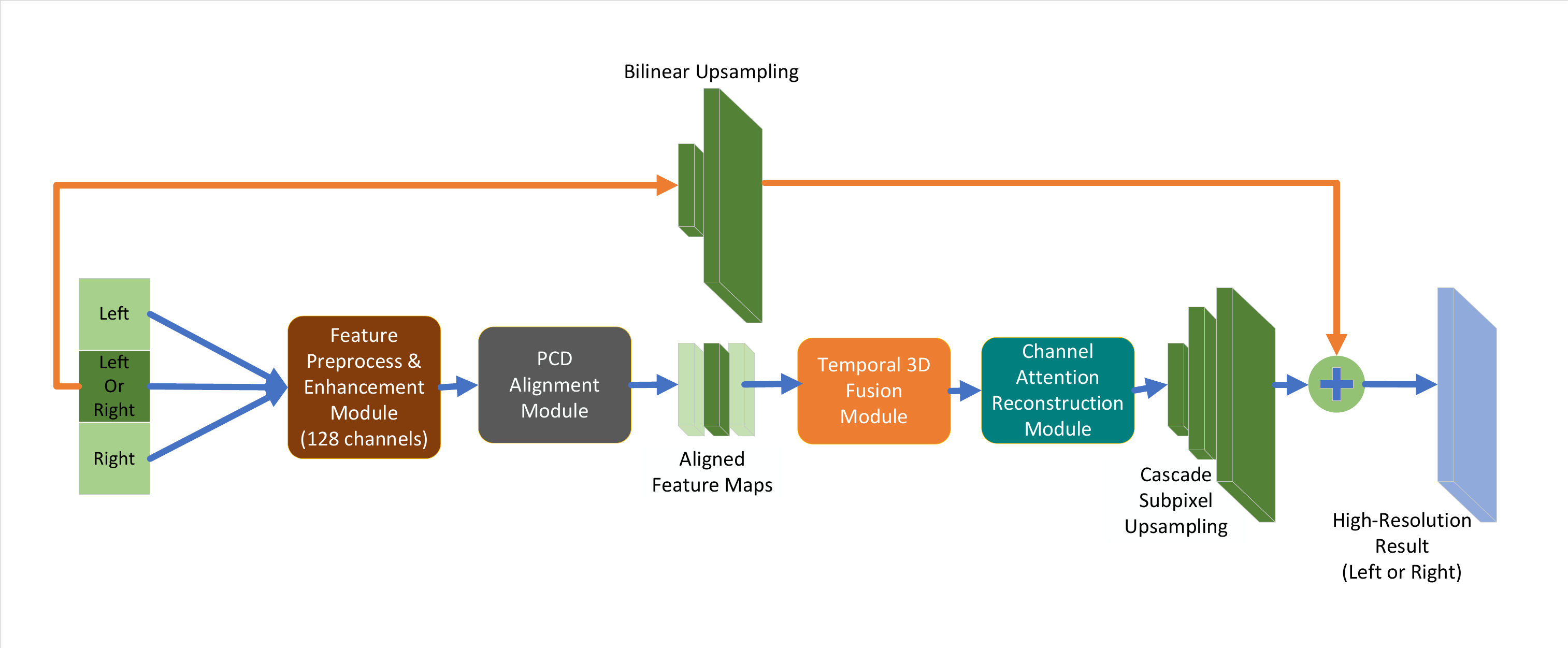}
	    \caption{The Modern\_SR Team: The network architecture of the proposed Stereo-EDVR.}
	    \label{Modern_SR2}
	\end{figure}
	
	\subsection{The Modern\_SR Team}
	The Modern\_SR team considered a pair of stereo images as two consecutive frames and developed a Stereo-EDVR network for SR. They aim at seeking a more general SR framework that can be used for different types of SR tasks. The network architecture of the proposed Stereo-EDVR is shown in Fig.~\ref{Modern_SR2}. First, a stereo image pair is formulated as a three-frame sequence by duplicating a left or right image. Then, an improved EDVR model with more channels is used to reconstruct an HR left or right image.

   \section{Acknowledgments}
We thank the NTIRE 2022 sponsors: Huawei, Reality Labs, Bending Spoons, MediaTek, OPPO, Oddity, Voyage81, ETH Zurich (Computer Vision Lab) and University of Wurzburg (CAIDAS).
    
	\section{Teams and Affiliations}
	\label{appendix}
	
	\subsection*{NTIRE2022 team}
	\noindent \textbf{\textit{Title:}} NTIRE 2022 Challenge on Stereo Image Super-Resolution
	
	\noindent \textbf{\textit{Members:}} {Yulan Guo}$^1$ (\textit{yulan.guo@nudt.edu.cn}), Longguang Wang$^1$, Yingqian Wang$^1$, Juncheng Li$^2$, Shuhang Gu$^3$, Radu Timofte$^4$
	
	\noindent \textbf{\textit{Affiliations:}} \\
	$^1$National University of Defense Technology\\
	$^2$The Chinese University of Hong Kong\\
	$^3$The University of Sydney\\
	$^4$University of W\"urzburg, ETH Z\"urich\\
	
	\subsection*{(1) The Fat, The Thin and The Young}
	\noindent \textbf{\textit{Title:}} Nonlinear Activation-Free Network
	
	\noindent \textbf{\textit{Members:}} Liangyu Chen$^1$ (\textit{chenliangyu@megvii.com}), Xiaojie Chu$^2$, Wenqing Yu$^1$
	
	\noindent \textbf{\textit{Affiliations:}} \\
	$^1$MEGVII Technology\\
	$^2$Peking University
	
	\subsection*{(2) BigoSR}
	\noindent \textbf{\textit{Title:}} SwiniPASSR: Swin Transformer based Parallax Attention Network for Stereo Image Super-Resolution
	
	\noindent \textbf{\textit{Members:}} Kai Jin$^1$ (\textit{jinkai@bigo.sg}), Zeqiang Wei$^2$, Sha Guo$^3$, Angulia Yang$^1$, Xiuzhuang Zhou$^4$, Guodong Guo$^5$
	
	\noindent \textbf{\textit{Affiliations:}} \\
	$^1$Bigo Technology Pte. Ltd.\\
	$^2$Smart Healthcare Innovation Lab, Beijing University of Posts and Telecommunications  \\
	$^3$Peking University\\
	$^4$School of Artificial Intelligence, Beijing University of Posts and Telecommunications \\ 
	$^5$Head of Institute of Deep Learning, Baidu Research
	
	\subsection*{(3) NUDT-CV\&CPLab}
	\noindent \textbf{\textit{Title:}} Stereo Image Super-Resolution Transformer
	
	\noindent \textbf{\textit{Members:}} Bin Dai$^1$ (\textit{daicver@gmail.com}), Feiyue Peng$^2$,  Huaxin Xiao$^1$, Shen Yan$^1$, Yuxiang Liu$^1$, Hanxiao Cai$^1$ 
	
	\noindent \textbf{\textit{Affiliations:}} \\
	$^1$College of Systems Engineering, National University of Defense Technology   \\
	$^2$College of Liberal Arts and Sciences, National University of Defense Technology \\
	
	\subsection*{(4) BUPT-PRIV}
	\noindent \textbf{\textit{Title:}} SwinIR-Impr
	
	\noindent \textbf{\textit{Members:}} Pu Cao (\textit{priv@bupt.edu.cn}), Yang Nie, Lu Yang, Qing Song
	
	\noindent \textbf{\textit{Affiliations:}} \\
	Pattern Recognition and Intelligent Vision Lab, Beijing University of Posts and Telecommunications
	
	\subsection*{(5) NKU\_caroline}
	\noindent \textbf{\textit{Title:}} PAMSwin
	
	\noindent \textbf{\textit{Members:}} Xiaotao Hu$^1$ (\textit{1979005820hux@gmail.com}), Jun Xu$^2$
	
	\noindent \textbf{\textit{Affiliations:}} \\
	$^1$College of Computer Science, Nankai University, Tianjin  \\
	$^2$School of Statistics and Data Science, Nankai University, Tianjin

	\subsection*{(6) BUAA-MC2}
	\noindent \textbf{\textit{Title:}} StereoSRT: A Stereo Image Super-Resolution Transformer
	
	\noindent \textbf{\textit{Members:}} Mai Xu (\textit{MaiXu@buaa.edu.cn}), Junpeng Jing, Xin Deng, Qunliang Xing, Minglang Qiao, Zhenyu Guan
	
	\noindent \textbf{\textit{Affiliations:}} \\
	Beihang University   \\

	\subsection*{(7) No War}
	\noindent \textbf{\textit{Title:}} Parallel Interactive Transformer for Stereo Image Super-Resolution
	
	\noindent \textbf{\textit{Members:}} Wenlong Guo (\textit{wlguo@zjut.edu.cn}), Chenxu Peng, Zan Chen
	
	\noindent \textbf{\textit{Affiliations:}} \\
	Zhejiang University of Technology
	
	\subsection*{(8) GDUT\_506}
	\noindent \textbf{\textit{Title:}} Parallax Res-Transformer Network
	
	\noindent \textbf{\textit{Members:}} Junyang Chen(\textit{3117002384@mail2.gdut.edu.cn}), Hao Li, Junbin Chen, Weijie Li, Zhijing Yang
	
	\noindent \textbf{\textit{Affiliations:}}\\
	Guangdong University of Technology
	
	\subsection*{(9) DSSR}
	\noindent \textbf{\textit{Title:}} Deformable Stereo Super-Resolution
	
	\noindent \textbf{\textit{Members:}} Gen Li (\textit{leegeun@yonsei.ac.kr}), Aijin Li, Lei Sun
	
	\noindent \textbf{\textit{Affiliations:}} \\
	Tencent OVBU\\

	\subsection*{(10) xiaozhazha}
	\noindent \textbf{\textit{Title:}} SwinFIR: Rethinking Image Restoration using Swin Transformer and Fast Fourier Convolution
	
	\noindent \textbf{\textit{Members:}} Dafeng Zhang (\textit{594112521@qq.com}), Shizhuo Liu
	
	\noindent \textbf{\textit{Affiliations:}} \\
	SRC-B \\
	
	\subsection*{(11) Zhang9678}
	\noindent \textbf{\textit{Title:}} Multi-stage Progressive Transformer for Stereo Image Super-Resolution
	
	\noindent \textbf{\textit{Members:}} Jiangtao Zhang (\textit{zjt9678@qq.com}), Yanyun Qu
	
	\noindent \textbf{\textit{Affiliations:}} \\
	Xiamen University \\
	
	\subsection*{(12) NTU607QCO-SSR}
	\noindent \textbf{\textit{Title:}} Transformer-based Super-Resolution with the Edge-aware Loss for Stereo Image Super-Resolution

	\noindent \textbf{\textit{Members:}} Hao-Hsiang Yang$^1$ (\textit{r05921014@ntu.edu.tw}), Zhi-Kai Huang$^1$, Wei-Ting Chen$^2$, Hua-En Chang$^1$, Sy-Yen Kuo$^1$
	
	\noindent \textbf{\textit{Affiliations:}} \\
	$^1$Department of Electrical Engineering, National Taiwan University  \\
	$^2$Graduate Institute of Electronics Engineering, National Taiwan University  \\

	\subsection*{(13) Supersmart}
	\noindent \textbf{\textit{Title:}} SwinRSTB
	
	\noindent \textbf{\textit{Members:}} Qiaohui Liang (\textit{qhliang1002@126.com})
	
	\noindent \textbf{\textit{Affiliations:}} \\
    Personal \\
    
    \subsection*{(14) LIMMC\_HNU}
	\noindent \textbf{\textit{Title:}} PAMSwinIR
	
	\noindent \textbf{\textit{Members:}} Jianxin Lin (\textit{linjianxin@hnu.edu.cn}), Yijun Wang, Lianying Yin, Rongju Zhang, Wei Zhao, Peng Xiao.
	
	\noindent \textbf{\textit{Affiliations:}} \\
    College of Computer Science and Electronic Engineering, Hunan University   \\
	
	\subsection*{(15) HIT-IIL}
	\noindent \textbf{\textit{Title:}} FFT Loss for Super-Resolution

	\noindent \textbf{\textit{Members:}} Rongjian Xu (\textit{1180301003@stu.hit.edu.cn}), Zhilu Zhang, Wangmeng Zuo
	
	\noindent \textbf{\textit{Affiliations:}} \\
	Harbin Institude of Technology  \\

	\subsection*{(16) HanSheng}
	\noindent \textbf{\textit{Title:}} Fine-tuned SwinIR

	\noindent \textbf{\textit{Members:}} Hansheng Guo$^{1}$ (\textit{hsguo.ai@gmail.com}), Guangwei Gao$^{2}$, Tieyong Zeng$^{1}$
	
	\noindent \textbf{\textit{Affiliations:}} \\
	$^{1}$The Chinese University of Hong Kong\\
	$^{2}$Nanjing University of Posts and Telecommunications\\

	\subsection*{(17) VIP-SSR}
	\noindent \textbf{\textit{Title:}} HFB-iPASSR (Hierarchical Feature Blended iPASSR)

	\noindent \textbf{\textit{Members:}} Joohyeok Kim$^1$ (\textit{kimjh4273@unist.ac.kr}), HyeonA Kim$^2$, Eunpil Park$^1$, Jae-Young Sim$^{1,2}$.
	
	\noindent \textbf{\textit{Affiliations:}} \\
	$^{1}$Department of Electrical Engineering, Ulsan National Institute of Science and Technology\\
	$^{2}$Graduate School of Artificial Intelligence, Ulsan National Institute of Science and Technology

	\subsection*{(18) phc}
	\noindent \textbf{\textit{Title:}} Improved-PASSR

    \noindent \textbf{\textit{Members:}} Huicheng Pi (\textit{21126334@bjtu.edu.cn}), Shunli Zhang
	
	\noindent \textbf{\textit{Affiliations:}} \\
	Beijing Jiaotong University  \\

	\subsection*{(19) qylen}
	\noindent \textbf{\textit{Title:}} iPASSR-Transformer Net

	\noindent \textbf{\textit{Members:}} Jucai Zhai (\textit{jucaizhai@stu.pku.edu.com}), Pengcheng Zeng, Yang Liu, Chihao Ma.
	
	\noindent \textbf{\textit{Affiliations:}} \\
	Peking University  \\
	
	\subsection*{(20) Modern\_SR}
	\noindent \textbf{\textit{Title:}} Stereo-EDVR

	\noindent \textbf{\textit{Members:}} Yulin Huang$^1$ (\textit{815018345@qq.com}), Junying Chen$^2$
	
	\noindent \textbf{\textit{Affiliations:}} \\
	$^1$City University of Hong Kong\\
	$^2$South China University of Technology  \\
	

	

{\small
\bibliographystyle{unsrt}
\bibliographystyle{ieee_fullname}

}

\end{document}